%% file: paper.tex
\documentclass[conference]{IEEEtran}
\usepackage{cite}
\usepackage{amsmath,amssymb,amsfonts}
\usepackage{graphicx}
\usepackage{multirow}
\usepackage{textcomp}
\usepackage{xcolor}
\ifCLASSOPTIONcompsoc \usepackage[caption=false,font=normalsize,labelfon
t=sf,textfont=sf]{subfig}
\else
\usepackage[caption=false,font=footnotesize]{subfig}
\fi

\usepackage{amsmath}
\usepackage{mathtools}

\usepackage[ruled,lined, noend,linesnumbered]{algorithm2e}
\SetKwIF{If}{ElseIf}{Else}{if}{}{else if}{else}{end if}%
\SetKwFor{While}{while}{}{end while}%
\SetKwFor{For}{Train for}{}{end for}%
\SetKwRepeat{Do}{do}{while}

\usepackage{algorithmicx}
\usepackage{textcomp}
\usepackage{filecontents}
\usepackage{microtype}
\usepackage{float}
\usepackage{adjustbox}
\usepackage{booktabs,makecell,tabularx}
\usepackage{url}
\usepackage{booktabs}
\usepackage{hyperref}

\newcolumntype{C}[1]{>{\centering\arraybackslash}p{#1}}
\newcolumntype{L}{>{\raggedright\arraybackslash}X}
\usepackage{siunitx}
\usepackage[utf8]{inputenc}

\usepackage{etoolbox}
\usepackage{xparse}

\graphicspath{{./}}

\def\BibTeX{{\rm B\kern-.05em{\sc i\kern-.025em b}\kern-.08em
    T\kern-.1667em\lower.7ex\hbox{E}\kern-.125emX}}
\begin{document}

\title{GraVAC: Adaptive Compression for Communication-Efficient Distributed DL Training}

\author{\IEEEauthorblockN{Sahil Tyagi}
\IEEEauthorblockA{\textit{Indiana University Bloomington, USA} \\
styagi@iu.edu}
\and
\IEEEauthorblockN{Martin Swany}
\IEEEauthorblockA{\textit{Indiana University Bloomington, USA} \\
swany@iu.edu}
}

\maketitle

\begin{abstract}
	\input{abstract}
\end{abstract}



\input{intro}

\input{bg}

\input{design}

\input{eval}

\input{conclusion}

\end{document}

%% file: abstract.tex
Distributed data-parallel (DDP) training improves overall application throughput as multiple devices train on a subset of data and aggregate updates to produce a globally shared model.
The periodic synchronization at each iteration incurs considerable overhead, exacerbated by the increasing size and complexity of state-of-the-art neural networks.
Although many gradient compression techniques propose to reduce communication cost, the ideal compression factor that leads to maximum speedup or minimum data exchange remains an open-ended problem since it varies with the quality of compression, model size and structure, hardware, network topology and bandwidth.
We propose \emph{GraVAC}, a framework to dynamically adjust compression factor throughout training by evaluating model progress and assessing gradient information loss associated with compression.
\emph{GraVAC} works in an online, black-box manner without any prior assumptions about a model or its hyperparameters, while achieving the same or better accuracy than dense SGD (i.e., no compression) in the same number of iterations/epochs.
As opposed to using a static compression factor, \emph{GraVAC} reduces end-to-end training time for ResNet101, VGG16 and LSTM by 4.32$\times$, 1.95$\times$ and 6.67$\times$ respectively.
Compared to other adaptive schemes, our framework provides 1.94$\times$ to 5.63$\times$ overall speedup.

%% file: intro.tex
\section{Introduction}\label{sec:intro}

Deep Learning (DL) is a supervised machine learning approach that optimizes a loss function over a non-convex surface by comparing model predictions with ground truth.
Each training iteration in DL involves forward and backward pass, i.e., generate predictions from input data, assess loss, compute gradients and update model parameters via optimization method like gradient descent.
Training is an iterative process, typically involving multiple passes over the entire dataset where each pass is called an \emph{epoch}.
DL is also heavily influenced by certain \emph{hyperparameters} that affect training speed, quality, or both.
Commonly used hyperparameters are learning rate, momentum, batch size, weight decay, epochs, activation function, etc.

Distributed data-parallel (DDP) methods further scale training across multiple nodes that train a globally shared model with I.I.D. data (independent and identically distributed) by periodically aggregating locally computed gradients at the end of each iteration.
The compute requirements to train DL models doubles every 3.5 months \cite{b0}, while the compute gains in chip design for ML accelerators and bandwidth gains in telecommunications networks double every 24 and 18 months \cite{b1, b2}.
Thus, the infrastructure required to train state-of-the-art models tends to fall behind their compute and networking demands.
Since upgrading network stack in the cloud, datacenter and HPC clusters can be infrequent as compared to appending new accelerators in pre-existing systems, gradient communication tends to be the major bottleneck in distributed training \cite{b3}.

Different compression techniques have been proposed in recent years to mitigate this synchronization overhead.
However, the optimal compression factor (CF) that minimizes data exchange or end-to-end training time depends on the model itself (i.e., its size, structure and depth), available network bandwidth and the compression overhead itself.
Unlike traditional HPC and distributed computing applications that only measure parallel efficiency, DDP training has an additional statistical efficiency associated with it.
Although the amount of computation performed on each iteration is the same, some iterations tend to be more crucial than others towards the overall learning of the model.
Updates are especially sensitive in early stages and to hyperparameters like learning rate schedule, momentum and weight decay \cite{b4}.
It would thus be intuitive to compare information loss in gradients on account of compression, and use a lower CF when considerably more information is lost and a higher CF when most information is preserved under compression.
We can subsequently increase compression as training continues and gradients saturate, and decrease it back during the aforementioned critical stages.

We take into account the parallel and statistical efficiency aspect of gradient compression in this work: a high CF improves overall throughput (i.e., number of samples processed per second) by reducing communication cost, but increases information loss in the gradients resulting in either slower or insignificant updates.
The two metrics in DDP compression are pareto-related as one improves at the detriment of the other.
We propose \textit{\textbf{GraVAC}}: \textbf{\{Gra\}}dient \textbf{\{V\}}ariance-based \textbf{\{A\}}daptive \textbf{\{C\}}ompression \footnote{Code available at \href{https://github.com/sahiltyagi4/GraVAC}{https://github.com/sahiltyagi4/GraVAC}} to dynamically adjust CF by comparing information loss from compression with that of the original gradients computed in backpropagation.
\emph{GraVAC} evaluates different CFs in a given search space and determines the CF that best balances parallel and statistical efficiency in DDP training with compression.
We validate our approach over a variety of DL models and directly compare with static CF on compressors like Top-$\mathit{k}$ \cite{b5}, Deep Gradient Compression or DGC \cite{b6}, Redsync \cite{b8} and Random-$\mathit{k}$ \cite{b5}.

%% file: bg.tex
\section{Background and related work}\label{sec:bg}

DDP training can be implemented either via MPI-based collectives (AllReduce) \cite{b9, b10, b11} or using one or more centralized parameter servers (PS) \cite{b12} to accumulate and distribute model updates among workers.

\subsection{Scaling Efficiency of DDP Training}\label{subsec:scalingddp}

DL training is an iterative process that involves parameter updates at each step via gradient descent (GD) \cite{b13}.
Full GD uses entire training data at every step, making the whole process slow and compute-intensive, while Stochastic GD processes a single sample and does not vectorize multiple samples on fast accelerators.
Mini-batch GD is the optimal middle ground between Full and Stochastic GD where \emph{b} samples are randomly sampled from I.I.D. data.
Eqn. (\ref{eqn:distsgd}) describes the update rule in mini-batch GD where parameters $\mathit{w}$ at $(\mathit{i}+1)$-\textit{th} iteration on $\mathit{N}$ workers minimize loss function $\mathit{\mathcal{L}(\cdot)}$ on input samples $\mathit{x_{j}}$ of size $\mathit{b}$ from distribution $\mathcal{X}_{j}$ with learning rate $\mathit{\eta}$.
With weak scaling, we can increase the amount of per-iteration work by adding more workers and keeping per-worker batch-size $\mathit{b}$ the same.

\begin{equation}
w_{i+1} = w_{i} - \eta \dfrac{1}{N} \sum_{n=1}^{n=N}{\dfrac{1}{|b|} \sum_{j \in b} \dfrac{\partial}{\partial w_{i}} \mathcal{L}(x_{(j,n)},w_{i})}
\label{eqn:distsgd}
\end{equation}

The \emph{ideal} throughput of a distributed application $\mathit{T_{N}}$ executed across $N$ workers is $N$ times the throughput of a single worker $\mathit{T_{1}}$.
The deviation is measured via ``scaling efficiency`` in Eqn. \ref{eqn:scaleout}.
Assuming negligible IO overhead, iteration time in dense SGD is bounded by computation and communication time (Eqn. (\ref{eqn:itrtime})).
It may be possible to overlap communication with computation, but only partially since the latter is comparatively much lower on modern GPUs and TPUs.
Model communication has been shown to be an order of hundreds or even thousands of magnitudes higher than gradient computation.
Thus, frequent synchronization ($t_{sync}$) is the bottleneck that halts linear scaling in DDP.
Table \ref{table:models} describes the size, density and convergence target of ResNet101 \cite{b14}, VGG16 \cite{b15} and LSTM \cite{b16} with dense SGD communication.
Latency is further exacerbated on constrained networks with limited bandwidth as large volumes of data is exchanged by multiple workers simultaneously.

\begin{subequations}
	\begin{equation}
		\eta_{scaling} = T_{N}/N\cdot T_{1}
		\label{eqn:scaleout}
	\end{equation}
	\begin{equation}
		t_{iter} \approx t_{compute} + t_{sync}
		\label{eqn:itrtime}
	\end{equation}
\end{subequations}

\begin{table}[!t]
\renewcommand{\arraystretch}{1.3}
	\caption{DL model description}
	\centering
	\begin{tabular}{|c|c|c|c|c|}
		\hline
		\bfseries Model & \bfseries Layers & \bfseries Size (MB) & \bfseries Dataset & \bfseries Test target \\
		\hline
		ResNet101 & 101 & 170 & CIFAR10 & 80\% Top-1 \\
		\hline
		LSTM & 2 & 252 & PTB & 22.0 PPL \\
		\hline
		VGG16 & 16 & 528 & CIFAR100 & 90\% Top-5 \\
		\hline
	\end{tabular}
\label{table:models}
\end{table}

\begin{figure}
\hspace*{-0.5cm}
\subfloat[Scaling efficiency in DDP]{\includegraphics[width=0.25\textwidth]{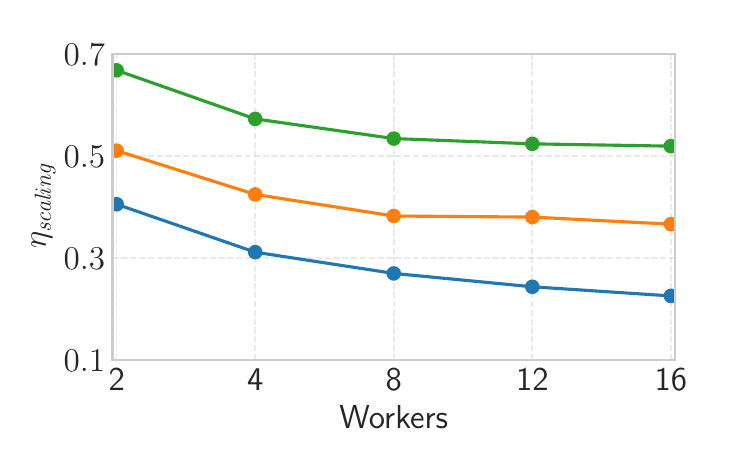}
\label{fig:scaleoutfactor}}
\subfloat[Initial gradient sensitivity]{\includegraphics[width=0.25\textwidth]{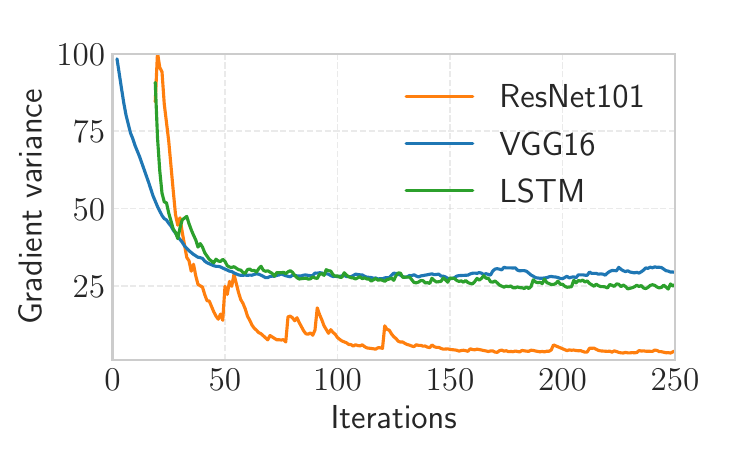}
\label{fig:allcritical}}
\caption{Communication overhead and early critical period in DDP training.} \label{fig:bgplots}
\end{figure}

For a DL model with a total of $M$ parameters, the time cost based on the $\alpha$-$\beta$ communication model (where $\alpha$ is the latency and $\beta$ is the inverse of bandwidth) for tree-based allreduce is $(2\alpha logN + 2MlogN \beta)$ \cite{b17}.
For ring-based allreduce, this becomes $2(N - 1)\alpha + 2M\beta (N-1)/N$.
Hence, communication cost increases as more workers are added to the mix in distributed training.
Fig. \ref{fig:scaleoutfactor} shows how overall throughput deviates from the ideal as cluster-size increases.
The scaling efficiency is also influenced by the message size, i.e., total gradients/parameters to be communicated.
In dense SGD, we observed scaling to be affected by the tensor-size distributions across the layers of a model as well.
For e.g., LSTM has a better $\eta_{scaling}$ than ResNet101 despite being a larger model.
This is because parameters in LSTM are spread across just 2 layers, compared to 101 in ResNet101.


\subsection{Gradient Variance in Deep Learning}\label{subsec:gradvar}

Prior work has demonstrated that gradient information can help measure the statistical efficiency of distributed training \cite{b18, b19}.
There is a strong correlation between changes in the eigen values of second-order hessian \cite{b20} and first-order gradients (i.e., variance).
\cite{b21, b22} explores how gradients behave in early stages of DL training and during certain critical periods, influenced by hyperparameters like learning rate schedule, gradient clipping and type of SGD used (e.g., zero, first or second-order moments).
Fig. \ref{fig:allcritical} attests those findings where we plot variance over the starting iterations and notice how drastically the gradients change and saturate over training.

\subsection{Gradient Compression}\label{subsec:sotacompress}

Many lossy compression techniques have been proposed for DDP and federated learning in recent years.
Lossy compression incurs a fundamental trade-off between data-size and information loss; one can either reduce message size by losing more information, or preserve data quality by keeping majority of the original bits intact.
In the context of DDP, higher CF reduces communication time at the cost of accuracy degradation or more steps/epochs required for the same convergence.
CF measures the size of original gradients to the size of compressed tensors.
E.g., compressing 10\% gradients gives CF of 10$\mathsf{x}$, while 1\% gives 100$\mathsf{x}$.
Lossy compression can be broadly classified into \emph{quantization}, \emph{sparsification} or \emph{low-rank approximations}.

The bit-width of single-precision (32-bit) floats is reduced in gradient quantization.
Techniques like automatic mixed precision (AMP) \cite{b23} reduces gradients to half-precision, resulting in 2$\mathsf{x}$ CF.
QSGD \cite{b24} balances the trade-off between accuracy and quantization precision.
1-bit SGD \cite{b25} reduces 32-bit floats to 1-bit and propagates quantization error via error-feedback.
Sparsification methods communicate only a fraction of the gradient values along with their indices and set everything else to 0.
Top-\textit{k} sparisifies by extracting the top \textit{k\%} values while Random-\textit{k} does so randomly with negligible compression overhead.
DGC discards gradients below a certain threshold along with using momentum correction and gradient clipping.
Methods like Redsync \cite{b38} combine quantization and sparsification, but the estimation quality is not accurate \cite{b26}.
Approaches like PowerSGD \cite{b27} and Pufferfish \cite{b28} achieve compression via low-rank updates.
The former can be viewed as adding regularization in DL, while the latter performs low-rank factorization on fully connected, convolutional and LSTM layers.

\vspace{0.1cm}
\emph{What should be the ideal CF in Compression-based DDP?}

\begin{figure}
\hspace*{-0.5cm}
\subfloat[Top-\textit{k}]{\includegraphics[width=0.25\textwidth]{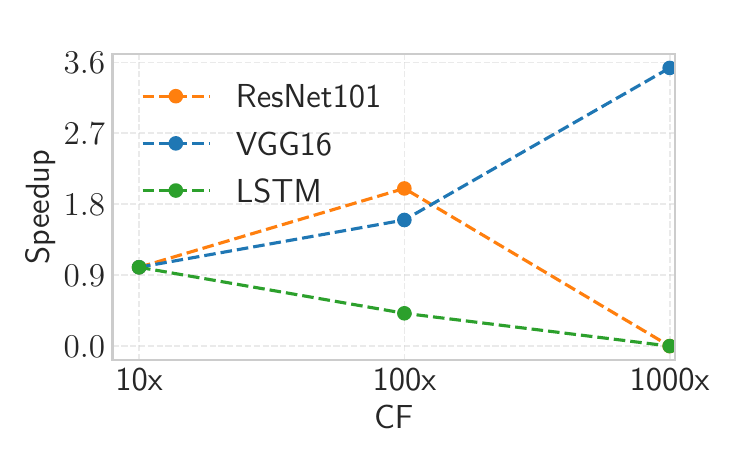}
\label{fig:topkbaseline}}
\subfloat[Random-\textit{k}]{\includegraphics[width=0.25\textwidth]{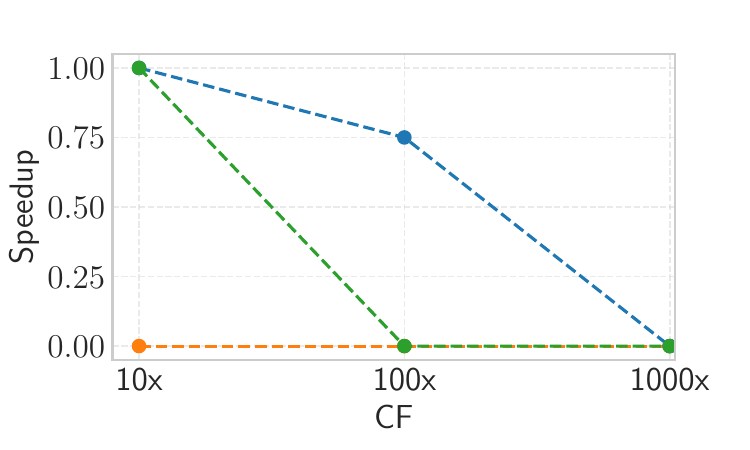}
\label{fig:randomkbaseline}}
\caption{CF with maximal speedup (to reach Table \ref{table:models} targets) varies for each model and compression technique used.
The results are normalized by 10$\mathsf{x}$ CF while a speedup of 0.0 implies convergence failure.} \label{fig:compressionbaseline}
\end{figure}

\vspace{0.1cm}
The ideal CF is one that reduces communication time without trimming too much gradients which can be detrimental to final model.
Compression has its own associated costs depending on the target CF and computational complexity of the mechanism itself.
These factors affect both the parallel efficiency of distributed training as well as statistical inefficiency due to information loss from compression.
Fig. \ref{fig:compressionbaseline} aptly demonstrates this where the CF that gives maximum speedup varies for each model and compression technique employed.
The models are trained to Table \ref{table:models} targets.
ResNet101 on Top-\textit{k} achieves most speedup at 100$\mathsf{x}$, while VGG16 and LSTM peak at CFs 1000$\mathsf{x}$ and 10$\mathsf{x}$ respectively.
On the other hand, ResNet101 fails to converge for any CF with Random-\textit{k} compression.
VGG16 and LSTM converged with 10$\mathsf{x}$ and failed with other CFs.
Although a typical ML practitioner may not necessarily need to think about a plethora of compression methods, choosing the right CF with any compressor and DL model that minimizes training time, or even converges successfully, presents a non-trivial challenge.

Dynamic compression mechanisms like AdaQS \cite{b29} perform quantization using gradient mean to standard deviation ratio (MSDR).
Systems like Accordion \cite{b30} and ScaDLES \cite{b300} switch between low and high compression based on critical regime identification.
We tackle the ideal CF exploration problem in \emph{GraVAC} in a gradient-driven manner by comparing variance of prior and post-compression gradients.
For clarity, prior-compression gradients refer to the original tensors computed in backward pass.
By measuring the information lost in compression, we dynamically adjust CF over each iteration.
Starting with a low CF initially, we gradually increase compression as training progresses.
On encountering senstive or critical regions, \emph{GraVAC} switches to a lower CF that least degrades convergence.

%% file: design.tex
\section{Design and implementation}\label{sec:design}

In this section, we first describe the trade-off between parallel and statistical efficiency of DDP training \emph{with} compression.
Then we describe the metrics ``compression gain`` and ``compression throughput`` to combine the two, and explain \emph{GraVAC}'s adaptive compression algorithm.

\subsection{Parallel Efficiency of Gradient Compression}\label{subsec:paraleff}

The end goal of gradient compression is to improve DDP scaling efficiency.
Application scaling is governed by the DDP mechanism (ring-based, tree-based allreduce or parameter servers), communication library used (MPI, NCCL \cite{b10}, Gloo \cite{b9} or RPC) and available bandwidth.
Keeping the latter and network infrastructure aside, speedup in any DL model depends on the target CF, quality of estimation and compression overhead.
The overall iteration time in Eqn. \ref{eqn:itrtime} is adjusted for compression as $$\mathit{t_{iter}^{(\mathit{c})}} \approx \mathit{t_{compute}} + \mathit{t_{sync}^{(\mathit{c})}} + \mathit{t_{compress}^{(\mathit{c})}} + \mathit{t_{decompress}^{(\mathit{c})}}$$
where it takes $\mathit{t_{compress}^{(\mathit{c})}}$ time to reduce gradients to CF $\mathit{c}$ such that it reduces communication time to $\mathit{t_{sync}^{(c)}}$.
$\mathit{t_{decompress}^{(\mathit{c})}}$ is the time taken to reconstruct the compressed gradients to the same dimension as the original gradients.
A viable compressor must have its compression time considerably lower than synchronization time.

The parallel efficiency of a distributed application suffers with more workers due to higher synchronization costs.
Improving the network bandwidth alleviates this to only a certain extent.
\cite{b3} investigates how DDP throughput improves marginally with higher bandwidth.
They observed that ResNet50 peaks to 75\% scale-out on a 25 Gbps network and remains the same even for 100 Gbps.
Its because network transport implementation of current DL frameworks cannot fully utilize the available network bandwidth.
Thus, even though cloud providers like GCP provide anywhere from 10-32 Gbps bandwidth depending on the machine type and VM size, they may not be utilized to their full potential.

\begin{figure}
\hspace*{-0.5cm}
\subfloat[Relative throughput]{\includegraphics[width=0.25\textwidth]{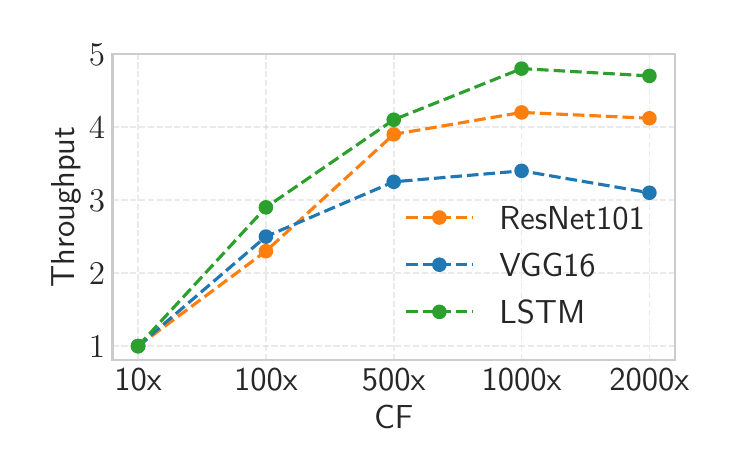}
\label{fig:tputCF}}
\subfloat[Relative communication]{\includegraphics[width=0.25\textwidth]{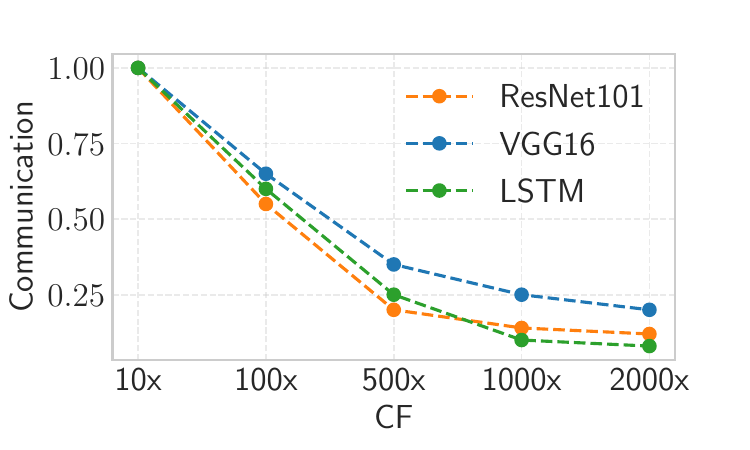}
\label{fig:commCF}}
\caption{Throughput and communication speedup for layerwise DGC compression, normalized by 10$\mathsf{x}$ CF.} \label{fig:commtputCF}
\end{figure}

Fig. \ref{fig:commtputCF} shows how the throughput increases and communication overhead reduces with compression.
The results are relative to CF 10$\mathsf{x}$ for each model.
We perform layerwise DGC compression over a 32 GPU cluster.
System throughput is determined only by compression overhead and communication time as the compute time in backpropagation stays the same across all CFs.
Based on the compressor used, compression latency may vary with target CF.
For e.g., it decreases with larger CF as Top-\textit{k} uses max-heap and sorts the top \textit{k}\% elements in $O(N + \mathit{k}\log{}\mathit{k})$ time.
Throughput for ResNet101 and VGG16 saturates at 500$\mathsf{x}$ and does not improve thereafter, while LSTM saturates at 1000$\mathsf{x}$ (Fig. \ref{fig:tputCF}).
Communication savings also diminish at higher CFs due to small message size and network saturation (Fig. \ref{fig:commCF}).
Thus, the highest CF may not necessarily correspond to the largest throughput.


\subsection{Statistical Inefficiency of Gradient Compression}\label{subsec:stateff}

Gradient compression mechanisms rely on \emph{error-feedback} \cite{b33, b34} which essentially acts as delayed updates, as commonly noted in asynchronous training.
The gradients ineligible for compression in the current iteration are not discarded, but added to \textit{residual gradients} which in turn are added to gradients computed in the next iteration.
Residual gradients and error-feedback helps preserve important features and is critical to convergence \cite{b5,b6,b007}.
Applying compression without error-feedback has been shown to achieve lower accuracy in deep learning models \cite{b33}.
At the same time, residual gradients can sometimes degrade generalization performance due to stale updates.

\begin{figure*}
\hspace*{-0.3cm}
\subfloat[10$\mathsf{x}$ CF]{\includegraphics[width=0.25\textwidth]{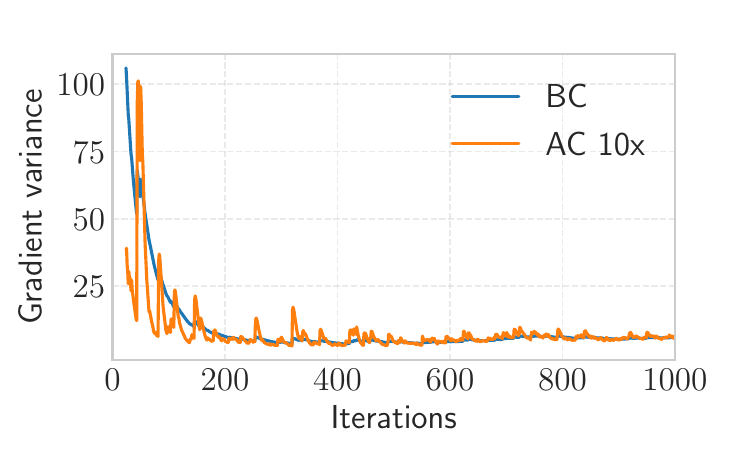}
\label{fig:resnet10xgnorm}}
\subfloat[1000$\mathsf{x}$ CF]{\includegraphics[width=0.25\textwidth]{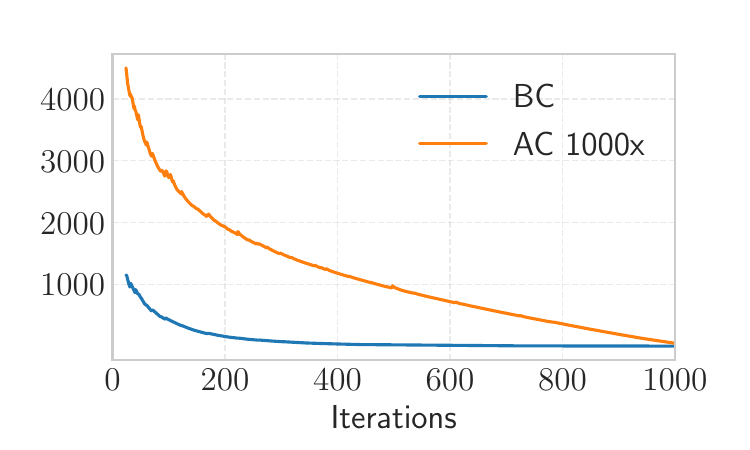}
\label{fig:resnet1000xgnorm}}
\subfloat[Convergence curve]{\includegraphics[width=0.25\textwidth]{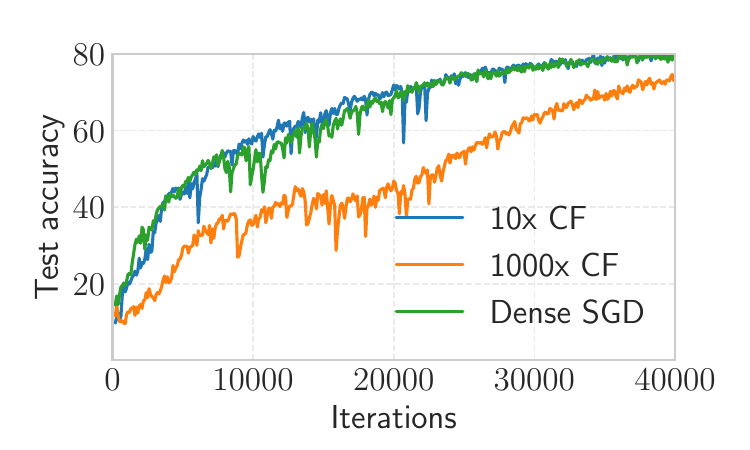}
\label{fig:resnet10gnormacc}}
\subfloat[Compression gain]{\includegraphics[width=0.25\textwidth]{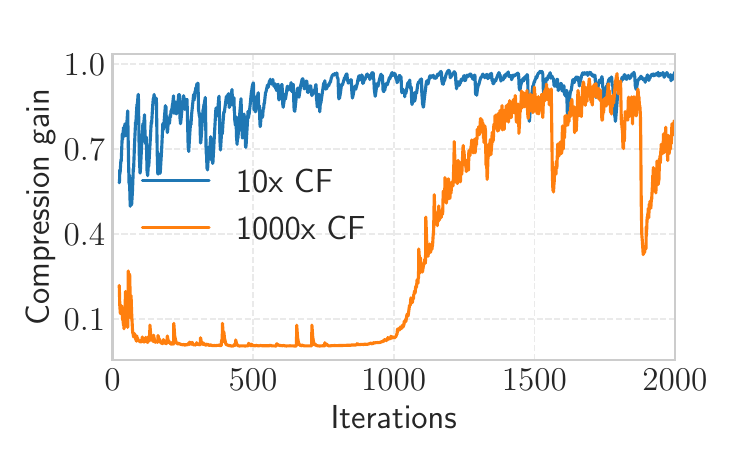}
\label{fig:resnetcompgain}}
\caption{ResNet101: Prior and Post-Compression gradients, test accuracy and compression gain for CFs 10$\mathsf{x}$ and 1000$\mathsf{x}$.} \label{fig:resnetpriorpostcomp}
\end{figure*}

\begin{figure*}
\hspace*{-0.3cm}
\subfloat[10$\mathsf{x}$ CF]{\includegraphics[width=0.25\textwidth]{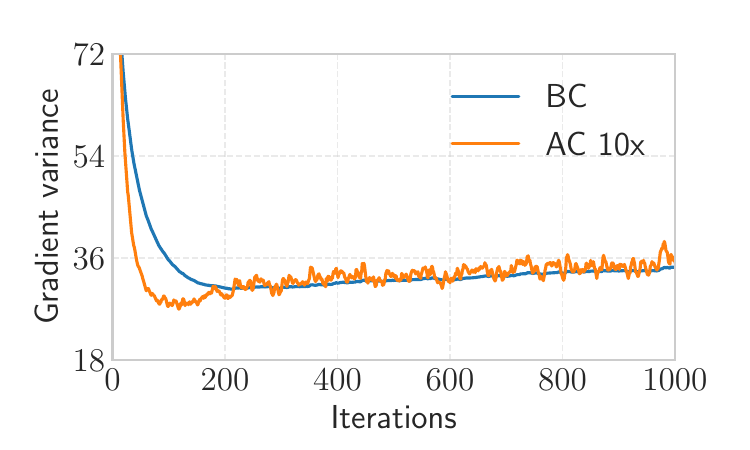}
\label{fig:vgg10xgnorm}}
\subfloat[1000$\mathsf{x}$ CF]{\includegraphics[width=0.25\textwidth]{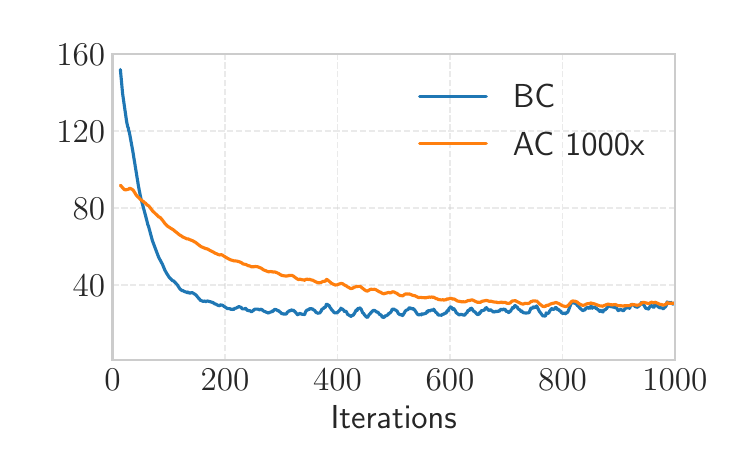}
\label{fig:vgg1000xgnorm}}
\subfloat[Convergence curve]{\includegraphics[width=0.25\textwidth]{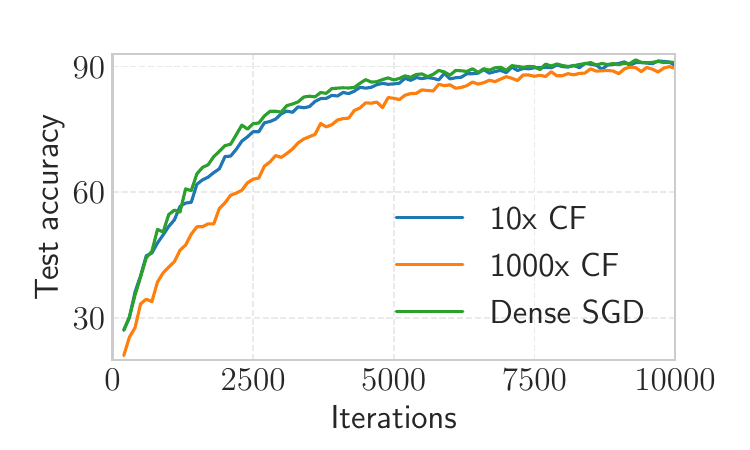}
\label{fig:vgg10gnormacc}}
\subfloat[Compression gain]{\includegraphics[width=0.25\textwidth]{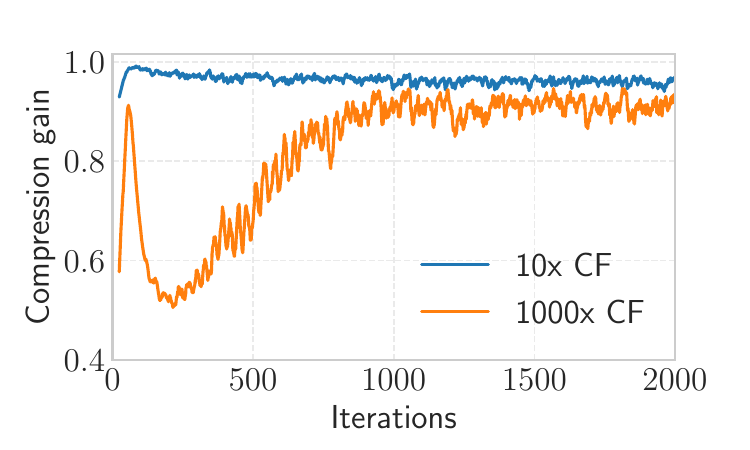}
\label{fig:vggcompgain}}
\caption{VGG16: Prior and Post-Compression gradients, test accuracy and compression gain for CFs 10$\mathsf{x}$ and 1000$\mathsf{x}$.} \label{fig:vggpriorpostcomp}
\end{figure*}

\begin{figure*}
\hspace*{-0.3cm}
\subfloat[10$\mathsf{x}$ CF]{\includegraphics[width=0.25\textwidth]{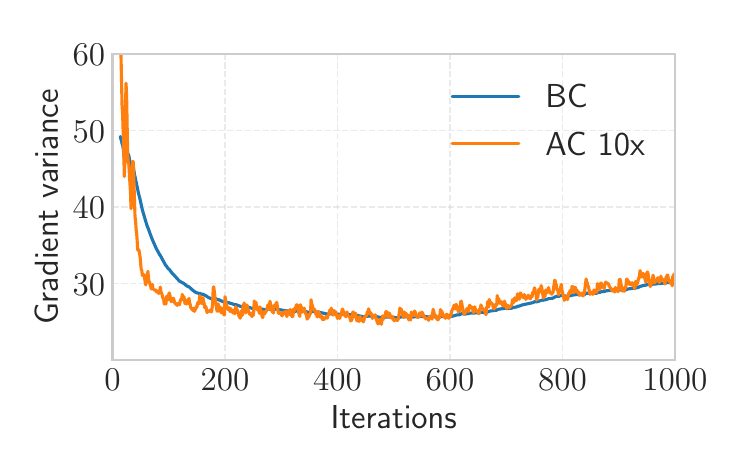}
\label{fig:lstm10xgnorm}}
\subfloat[1000$\mathsf{x}$ CF]{\includegraphics[width=0.25\textwidth]{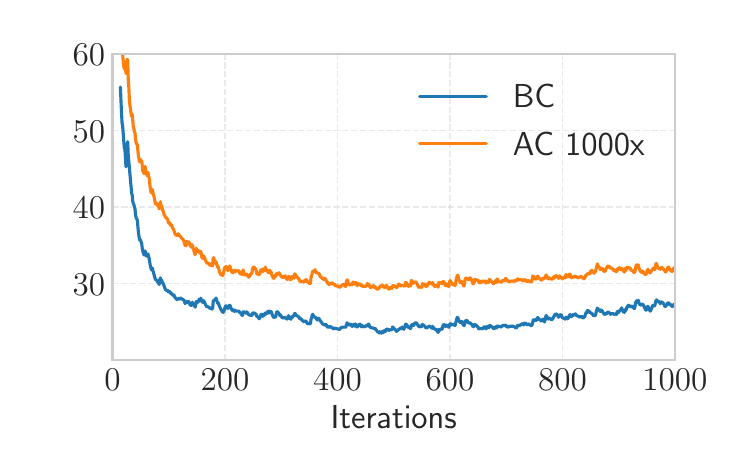}
\label{fig:lstm1000xgnorm}}
\subfloat[Convergence curve]{\includegraphics[width=0.25\textwidth]{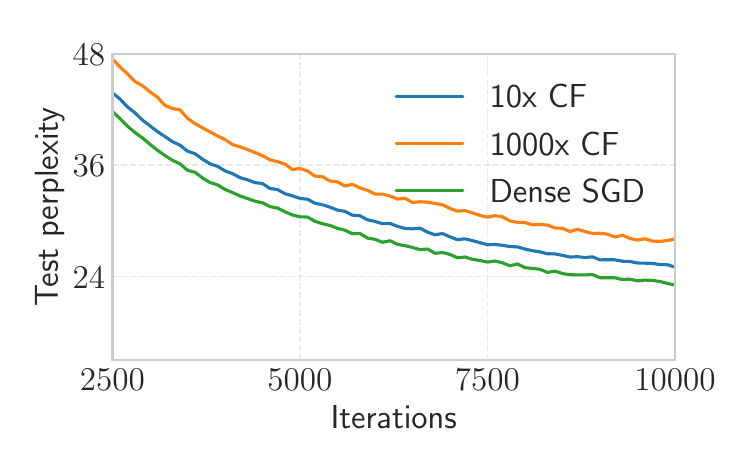}
\label{fig:lstmgnormacc}}
\subfloat[Compression gain]{\includegraphics[width=0.25\textwidth]{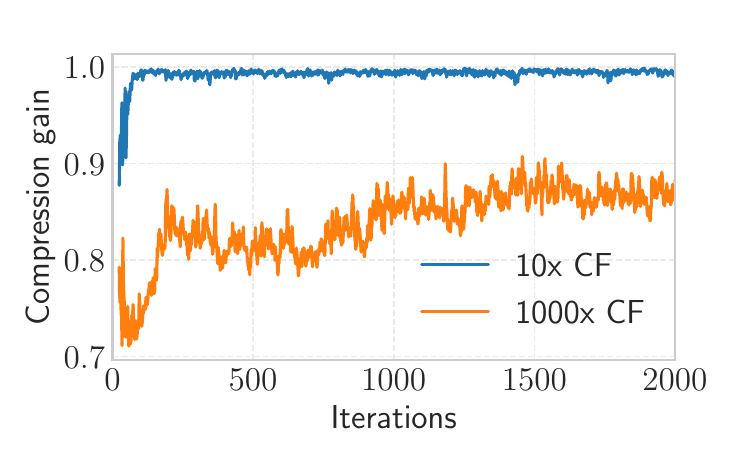}
\label{fig:lstmcompgain}}
\caption{LSTM: Prior and Post-Compression gradients, test perplexity (lower is better) and compression gain for CFs 10$\mathsf{x}$ and 1000$\mathsf{x}$.} \label{fig:lstmpriorpostcomp}
\end{figure*}

DDP training with very high CFs can negatively impact training time, convergence quality, or both if the compressed gradients are too sparse or quantized to update the model in any significant way.
\emph{It is thus crucial to have an indicator that quantifies information loss between compressed and the original gradients.}
We do so by comparing variance between the original and compressed tensors on every iteration and see how it relates to actual model convergence.
Denoting the original gradients as \emph{BC} (\textit{Before-Compression}) and compressed tensors as \emph{AC} (\textit{After-Compression}), we compare BC and AC tensors in two separate configurations with CFs 10$\mathsf{x}$ and 1000$\mathsf{x}$ in Fig. \ref{fig:resnetpriorpostcomp}, \ref{fig:vggpriorpostcomp} and \ref{fig:lstmpriorpostcomp}.
We compare the convergence curves for the two CFs with \emph{Dense SGD} (i.e., no compression) to see how much accuracy degrades with compression.

\emph{AC} 10$\mathsf{x}$ is nearly identical to its \emph{BC} counterpart in ResNet101 (Fig. \ref{fig:resnet10xgnorm}) while there is considerably more information loss in between \emph{BC} and \emph{AC} 1000$\mathsf{x}$ (Fig. \ref{fig:resnet1000xgnorm}).
This translates to their convergence curves in Fig. \ref{fig:resnet10gnormacc} as well where 10$\mathsf{x}$ and dense SGD runs follow a similar convergence trajectory while 1000$\mathsf{x}$ achieves considerably lower accuracy for the same iterations.

VGG16 follows a similar trend with 10$\mathsf{x}$ CF.
The \emph{BC} and \emph{AC} gradient variance (Fig. \ref{fig:vgg10xgnorm}) is nearly identical and so are the convergence curves for 10$\mathsf{x}$ and Dense SGD (Fig. \ref{fig:vgg10gnormacc}).
We notice a slight deviation between \emph{BC} and \emph{AC} at 1000$\mathsf{x}$ initially in Fig. \ref{fig:vgg1000xgnorm}, which correlates to slow convergence in the early iterations for 1000$\mathsf{x}$ in Fig. \ref{fig:vgg10gnormacc}.
As the deviation \emph{BC} and \emph{AC} decreases, we see both CFs converge to the same accuracy as Dense SGD in the same iterations.

The \emph{AC} 10$\mathsf{x}$ and 1000$\mathsf{x}$ gradients lie on similar scales as \emph{BC} in LSTM, although the higher CF has slightly higher variance (Fig. \ref{fig:lstm10xgnorm} and \ref{fig:vgg1000xgnorm}).
As seen from Fig. \ref{fig:vgg10gnormacc}, Dense SGD has the least perplexity (thus, better model quality), followed by 10$\mathsf{x}$ and 1000$\mathsf{x}$ CFs.

To compare the information loss between the original and gradients compressed to CF \emph{c}, we define a simplistic metric called \emph{Compression gain}.
As part of error feedback, we update the gradients such that $\mathit{g_{ef}^{(i)}} = \mathit{g_{0}^{(i)}} +\; \mathsf{residual\_gradients}^{(i-1)}$ for $i \ge 1$.
Here, $\mathit{g_{0}^{(i)}}$ are the original gradients calculated via backpropagation at iteration $i$, while $\mathsf{residual\_gradients}^{(i-1)}$ are left-overs from the last iteration $(i-1)$ and before, which are added back as part of error-feedback to produce $\mathit{g_{ef}^{(i)}}$ for the current iteration.
With compression operator $\mathcal{C}$, gradients are compressed as $\mathit{g_{c}^{(i)}}=\mathcal{C}[\mathit{g_{ef}^{(i)}}]$.
Compression gain is then measured as the ratio of expected variance of compressed gradients $\mathit{g_{c}^{(i)}}$ and the original gradients modified with error-feedback, i.e., $\mathit{g_{ef}^{(i)}}$: $$\mathsf{Compression\ gain} = \frac{\mathbb{E}[||g_{c}^{(i)}||^{2}]}{\mathbb{E}[||g_{ef}^{(i)}||^{2}]}$$

In prior work, gradient noise has been well studied in deep learning literature pertaining to divergence between locally-computed and aggregated gradients in DDP \cite{b19, b35, b36}.
These works use gradient information to tweak the global batch-size in DDP to optimize job completion time or allocate optimal resources for a job.
Instead of looking at local and global gradients, \emph{GraVAC}'s novelty comes from evaluating the noise between the original and compressed tensors.
The gradients computed over each iteration can be noisy.
Thus, we keep a moving average of the respective variances of the original and compressed gradients.
The computation and memory footprint of this approach is low since the window-size in moving average is finite and only a single-precision floating point is stored for every iteration.
Compression gain is bounded between $\{0,1]$ such that it is low when $\mathcal{C}$ trims too much information.
As models keep training, gradients saturate and higher compression becomes more viable in later stages of training.
Hence, compression gain increases over training as compressed tensors become more aligned with the original gradients.

We plot compression gains for the three models when training with fixed CF 10$\mathsf{x}$ and 1000$\mathsf{x}$ respectively, shown in Fig. \ref{fig:resnetcompgain}, \ref{fig:vggcompgain} and \ref{fig:lstmcompgain}.
In each model, 10$\mathsf{x}$ has higher compression gain than 1000$\mathsf{x}$ since more information is preserved in the smaller CF.
\emph{It should also be apparent that Dense SGD training has a constant gain of 1.0.}
For all models, convergence curve of 10$\mathsf{x}$ follows a similar trajectory as Dense SGD.
Correspondigly, the compression gain of 10$\mathsf{x}$ stays close to 1.0 throughout.
In ResNet101, gain of 1000$\mathsf{x}$ is low initially and grows in an oscillating manner, although still lower than gains of 10$\mathsf{x}$ and Dense SGD.
The low gains in the first 1000 iterations of CF 1000$\mathsf{x}$ correlates to the considerable gap between \emph{BC} and \emph{AC} gradients in Fig. \ref{fig:resnet1000xgnorm} and lower accuracy in Fig. \ref{fig:resnet10gnormacc}.
VGG16 is more robust to higher CFs (Fig. \ref{fig:vgg10gnormacc}), as also seen from the high compression gains of CF 1000$\mathsf{x}$ in Fig. \ref{fig:vggcompgain}.
For LSTM, compression gain for 10$\mathsf{x}$ stays close to 1.0 and between 0.8-0.9 for 1000$\mathsf{x}$.
The proximity of the two CFs to Dense SGD's gain of 1.0 is equivalent to their perplexity curves in Fig. \ref{fig:lstmgnormacc}.
From these results we see how compression gain serves as a viable indicator of the statistical efficiency of DDP with compression.

\subsection{Combining System Throughput and Compression Gain}\label{subsec:comptput}

As described earlier in \ref{subsec:sotacompress} as well as Fig. \ref{fig:compressionbaseline}, choosing a high CF unintuitively does not necessarily improve training time and may even degrade final model quality.
Thus, to account for both the parallel and statistical efficiency DDP training \emph{with} gradient compression, we combine \emph{system throughput} ($\text{T}_{system}$) and \emph{compression gain} into a single metric called \emph{Compression Throughput}: $$\text{T}_{compression} = \; \text{T}_{system} \; \times \; \mathsf{Compression\ gain}$$

If CF is high, system throughput would be high as well but compression gain would relatively be lower, decreasing the resulting $\text{T}_{compression}$.
On the other hand, compression gain will be high for a low CF, but system throughput will be lower due to relatively higher communication overhead.
\emph{With Compression Throughput, we capture this pareto-relationship between the parallel (system throughput) and statistical efficiency (compression gain) of gradient compression in DDP.}

\begin{algorithm}
	\SetKwProg{Pn}{procedure}{:}{\KwRet}
	\SetKwFunction{updatestep}{\text{UpdateStep}}
	\SetKwFunction{evalgravac}{\text{CheckGraVAC}}
	\caption{\emph{GraVAC}'s Adaptive Compression}\label{algo:adaptivecompression}
	\textbf{Input:} $\theta_{min}$, $\theta_{max}$, $\epsilon$, $\theta_{s}$, $\omega$, $\mathsf{window}$, compressor $\mathcal{C}$\\
		$w_{o} : $ initial model state, N: total nodes, b: per-worker batch-size, residual = 0; $\text{T}_{sys}, \text{T}_{compress}$ = empty()\\
		\For {i = 1,2,3...  \Comment{training iterations}} {
			$g_{o}^{(i)}, t_{o} = \nabla f(x^{(i)},w_{i})$
			\Comment{backpropagation}\\
			$g_{o}^{(i)} = g_{o}^{(i)} + $ residual
			\Comment{error-feedback}\\
			$g_{min}^{(i)}, t_{min} = \mathcal{C}(g_{o}^{(i)}, \theta_{min})$
			\Comment{compress to CF $\theta_{min}$}\\
			$\delta_{min} = \text{EWMA}$($\frac{||g_{min}^{(i)}||^{2}}{||g_{o}^{(i)}||^{2}}$)
			\Comment{$\theta_{min}$ compression gain}\\
			\BlankLine
				$g_{c}^{(i)}, t_{c}^{(i)} = \mathcal{C}(g_{min}^{(i)}, \theta_{s})$
				\Comment{compress to CF ($\theta_{s}\cdot \theta_{min}$)}\\
				$\delta_{c} = \text{EWMA}$($\frac{||g_{c}^{(i)}||^{2}}{||g_{o}^{(i)}||^{2}}$)
				\Comment{gain for CF ($\theta_{s}\cdot \theta_{min}$)}\\
				\BlankLine
				$t_{compress} = t_{min} + t_{c}$
				\Comment{total compression time}\\
				\BlankLine
				\If {$\delta_{c} \geq \epsilon:$} {
					$\tilde{g}^{(i)}$, $t_{s}$ = Aggregate($g_{c}^{(i)}$)
					\Comment{synchronize $g_{c}^{(i)}$}\\
					residual = $g_{o}^{(i)} - g_{c}^{(i)}$
					\Comment{update residual}\\
					$t_{iter}$ = $t_{o}$ + $t_{compress}$ + $t_{s}$
					\Comment{iteration time}\\
					\updatestep{$\theta_{s}\cdot \theta_{min}, \delta_{c}, t_{iter}$}					
					\BlankLine
				} \ElseIf {$\delta_{c} < \epsilon \; \text{and} \; \delta_{min} \geq \epsilon:$} {
					$\tilde{g}^{(i)}$, $t_{s}$ = Aggregate($g_{min}^{(i)}$)
					\Comment{synchronize $g_{min}^{(i)}$}\\
					residual = $g_{o}^{(i)} - g_{min}^{(i)}$
					\Comment{update residuals}\\
					$t_{iter}$ = $t_{o}$ + $t_{compress}$ + $t_{s}$
					\Comment{iteration time}\\
					\updatestep{$\theta_{min}, \delta_{min}, t_{iter}$}
					\BlankLine
				} \Else {
					$\tilde{g}^{(i)}$, $t_{s}$ = Aggregate($g_{o}^{(i)}$)
					\Comment{synchronize $g_{o}^{(i)}$}\\
					residual = 0
					\Comment{no residual gradients}\\
					$t_{iter}$ = $t_{o}$ + $t_{s}$
					\Comment{iteration time}\\
					\updatestep{$1, 1, t_{iter}$}
				}
				\BlankLine
				\BlankLine
				$w_{i+1} = w_{i} - \eta\cdot \tilde{g}^{(i)}$
				\Comment{apply SGD update}\\
				$\theta_{s}$ = \evalgravac{i, $\theta_{s}, \delta_{min}, \delta_{c}$}
		}		
		\BlankLine
		\BlankLine
		\BlankLine
		\Pn{\updatestep{$\theta, \delta, t_{iter}$}} {
			$\text{T}_{sys}$ = $\text{N}\cdot \text{b}/t_{iter}$
			\Comment{system throughput}\\
			$\text{T}_{compress}[\theta] = \text{T}_{sys}\cdot \delta$
			\Comment{compression throughput}
		}
		
		\BlankLine
		\BlankLine
		\BlankLine
		\Pn{\evalgravac{i, $\theta_{s}, \delta_{min}, \delta_{c}$}} {
			\If {i \% $\mathsf{window} == 0:$} {
				$\theta_{s} = \textsf{ScalingPolicy}(\theta_{s})$
				\Comment{compression scale-up}\\
				\BlankLine
				\If {$\omega \ge \frac{|\delta_{min} - \delta_{c}|}{\delta_{min}}:$} {
				$\theta_{min} = \theta_{s}\cdot \theta_{min}$				
				\Comment{scale-up minimum CF}			
				}
				\BlankLine
				\BlankLine
				ct = sort($\text{T}_{compress}.\text{values}()$)
				\Comment{$\text{T}_{compress}$ vals}\\
				\BlankLine
				\If {$|\frac{\text{ct}[-1] \;-\; \text{ct}[-2]}{\text{ct}[-2]}| \leq \omega:$} {
					\BlankLine
					$\theta_{ideal} = \text{T}_{compress}.get(\text{ct}[-2])$
					\Comment{ideal CF}\\
					return $\theta_{ideal}/\theta_{min}$		
					\Comment{gives optimal $\theta_{s}$}\\	
				} \Else {
					return $\theta_{s}$
					\Comment{else use old scaling factor}
				}
				\BlankLine
			}
		}
\end{algorithm}

We build \emph{GraVAC} as a modular extension on top of PyTorch's \cite{b31} DDP module \cite{b32} using Python in about 3000 lines of code.
A base $\mathsf{GravacOptimizer}$ wraps common SGD optimizers implemented in PyTorch by extending the base $\mathsf{torch.optim.Optimizer}$ class.
The optimizer takes an additional $\mathsf{Compressor}$ object that specifies the type of compression technique used.
We implement four pre-existing techniques as compressor classes in this paper: Top-\textit{k}, DGC, Redsync and Random-\textit{k}.
Compression for the appropriate CF and its gain is computed before the optimizer $\mathsf{step}$ function which applies the aggregated gradient updates on model parameters.

\vspace{0.15cm}
\emph{GraVAC Algorithm:} Alg. \ref{algo:adaptivecompression} describes \emph{GraVAC}'s approach of using compressor $\mathcal{C}$ to scale CFs in the exploration space [$\theta_{min}, \theta_{max}$], where each candidate CF is evaluated for $\mathsf{window}$ steps and incremented in step-size of $\theta_{s}$ w.r.t. $\theta_{min}$.
For e.g., scaling from CF 10$\mathsf{x}$ to CF 20$\mathsf{x}$ means $\theta_{s} = 20/10 = 2\mathsf{x}$.
The threshold $\epsilon$ denotes the minimum compression gain required for any CF to be eligible for communication in \emph{GraVAC}, while threshold $\omega$ is used to measure saturation in compression throughputs and for scaling up $\theta_{min}$.
We explain this in the following sections in more detail.
For every iteration, we compute gradients $g_{o}^{(i)}$ with model parameters $w_{i}$ on training sample $x^{(i)}$ in time $t_{o}$ (line 4).
To incorporate error-feedback,$\text{residual}$ holds the leftover gradients not communicated from previous iterations. 
The shape and memory size of tensors in $\text{residual}$ is the same as gradients itself.
As shown in line 5, we add residual gradients to the gradients computed in the current iteration.
In the first stage, we compress original gradients using $\mathcal{C}$ to compressed gradients $g_{min}^{(i)}$ corresponding to minimum CF $\theta_{min}$ (line 6).
We then compute the compression gain corresponding to $\theta_{min}$ (line 7), and smoothen out the inter-iteration gain through exponential weighted moving average (EWMA) smoothing.
In our evaluation, we set the EWMA smoothing factor to $N$/100, where $N$ is the number of participating workers.
We evaluate the next candidate CF by stepping up the previous $\theta_{min}$ and further compressing the already compressed gradients $g_{min}^{(i)}$ by stepsize $\theta_{s}$ (line 8).
Thus, candidate CF evaluated in this case is $\theta_{s}\cdot \theta_{min}$.
This is done as part of our multi-level compression strategy to avoid compressing the large, original tensors $g_{o}^{(i)}$ twice.
We measure the time savings of our multi-level approach in section \ref{subsec:multilevelgravac}.

Next, we compute the gradients and compression gain of candidate CF $\theta_{s}\cdot \theta_{min}$ (line 8-9), and denote the total compression time $t_{compress}$ as the sum of time to compress original gradients to $g_{min}^{(i)}$ (line 6) and the time to further compress $g_{min}^{(i)}$ to $g_{c}^{(i)}$ (line 8).
Based on the compression gains obtained and threshold $\epsilon$, we choose the appropriate gradients to call the collective operation on.
If the gain of our candidate CF meets $\epsilon$ (line 11), we go ahead and communicate compressed gradients $g_{c}^{(i)}$ among workers.
We update the residual gradients in accord with $g_{c}^{(i)}$ as well (line 13), calculate the total iteration time (line 14) and update the system as well as compression throughput for CF $\theta_{s}\cdot \theta_{min}$ via $\mathsf{UpdateS	tep}$ function.
$\text{T}_{compress}$ is a dictionary or a hashmap that stores compression throughput of each candidate CF, min-max CF as well as dense SGD setting (i.e., CF 1$\mathsf{x}$).

If the gain of $g_{c}^{(i)}$ does not meet the threshold, but gain $\delta_{min}$ of $\theta_{min}$ does (line 16), we instead synchronize compressed gradients $g_{min}^{(i)}$ corresponding to $\theta_{min}$.
In a similar fashion as before, we update the residuals, this time with $g_{min}^{(i)}$ instead of $g_{c}^{(i)}$ (line 18), compute iteration time and assess compression throughput.
\emph{It is important to remember that synchronization overhead to communicate $g_{min}^{(i)}$ is more than $g_{c}^{(i)}$ due to the former's lower CF. The trade-off we make in GraVAC is to incur higher communication latency for more accurate representation of the original gradients (measured by compression gain) and vice-versa.}

If both $\theta_{min}$ and currently evaluated CF do not meet the set threshold, we incur maximum communication latency by transmitting the original gradients via dense SGD (line 22).
In this case, residual gradients are set to 0 and no compression overhead is included as part of iteration time and computing system/compression throughput.
The CF and compression gain are both 1, as set in the $\mathsf{UpdateStep}$ function at line 25.

Following SGD update (line 26), we evaluate \emph{GraVAC} to assess the performance of CFs evaluated so far.
This happens at a frequency determined by $\mathsf{window}$.
Here, we adjust $\theta_{s}$ by a certain factor to scale up compression, determined by the chosen $\textsf{ScalingPolicy}$.
The scaling policy tunes compression only until the upper bound $\theta_{max}$.
We explore two scaling policies in this paper that we describe in detail under section \ref{subsec:adaptivegravac}.
After scaling $\theta_{s}$, we also assess if the minimum CF, i.e., $\theta_{min}$ can be scaled up as well.
The intuition is that as training progresses, model gradually starts converging as well and we can use higher compression even for the minimum CF later on.
In addition to candidate CFs, we thus scale up the minimum CF as well.
The transition is made if the current gain $\delta_{c}$ is within $\omega$\% of the gain of previous $\theta_{min}$ (line 34).
Once enough CFs are evaluated, we look at the two largest compression throughputs (line 36) and fetch the corresponding CF if they are within the bounds of $\omega$.
We do this as it means the compression throughput has saturated and thus, we pick the lower CF as $\theta_{ideal}$ (line 38) and send the appropriate step-size (line 39).
If the threshold $\omega$ is not met, we use $\theta_{s}$ as is.

\vspace{0.15cm}
\emph{When does compression scale-up?} As seen from Alg. \ref{algo:adaptivecompression}, the compression scale-up happens during \emph{GraVAC}'s evaluation phase where we scale the step-size $\theta_{s}$ in accordance with a specific scaling policy.
At the same time, we escalate the minimum CF $\theta_{min}$ to currently evaluated CF if the two compression gains are within $\omega$\% of each other.

\vspace{0.15cm}
\emph{When does compression scale-down?} Compression scale-down is determined by $\epsilon$ (shown via conditional statements lines 11-25).
If current CF loses considerably more information in compressed gradients $g_{c}^{(i)}$, we use the lower CF $\theta_{min}$.
If the latter fails to meet $\epsilon$ as well, we send uncompressed gradients $g_{o}^{(i)}$ as a last resort.

%% file: eval.tex
\section{Evaluation}\label{sec:eval}

\subsection{Cluster Setup and Training Hyperparameters}\label{subsec:cloudsetuphyper}

We evaluate \emph{GraVAC} on a 32 GPU setup on the Google Cloud Platform (GCP) across 8 VMs.
Each VM is a $\mathsf{n1}$-$\mathsf{standard}$-$\mathsf{8}$ machine type with 8 vCPUs, 30 GB system memory and 4 NVIDIA V100 GPUs with 16 GB VRAM each.
The machines are configured with PyTorch 1.10.1, CUDA 11.3, CUDA driver 465.19.01 and NCCL 2.10.3.

We evaluate the three models described in Table \ref{table:models}.
ResNet101 is trained with per-worker batch size 32, momentum 0.9, weight decay 0.0001 and SGD optimizer with initial learning rate (lr) 0.1 decayed by a factor of 10 at 9K and 14K iterations respectively.
VGG16 is also trained with per-worker batch-size 32, weight decay 0.0005, momentum 0.9 and SGD with fixed lr 0.1.
Lastly, LSTM is measured with test perplexity (i.e., exponential of test loss) with per-worker batch-size 20, momentum 0.9, weight decay 0.0001 and SGD with fixed lr 0.1.
The model is initialized with 1500 embedding dimensions and 2 hidden layers with 35 bptt steps.

We evaluate \emph{GraVAC} with different scaling policies and look at their convergence curves (i.e. test accuracy/perplexity vs. iterations), average compression throughput of candidate CFs and kernel density estimates (KDE) of training iterations using different CFs over the course of training.
KDE gives the distribution over the iterations for all CFs and plotted on the log-scale with smoothing bandwidth of $0.1$ passed to the gaussian KDE.

\subsection{\emph{GraVAC}'s Adaptive Compression Policies}\label{subsec:adaptivegravac}

In this section, we look at how \emph{GraVAC} achieves optimal CF for a given $\theta_{min}$, $\theta_{max}$, $\epsilon$, $\mathsf{window}$, $\omega$ and $\mathsf{stepsize}$.
To see how a model converges and communication costs vary by evaluating different candidate CFs in the search space, we employ an \emph{Exponential} policy that upscales CFs aggressively, and a relatively smoother \emph{Geometric} scaling policy that scales CFs as a geometric progression.
\vspace{0.1cm}
\subsubsection{Exponential scaling policy}

In this policy, we implement the $\textsf{ScalingPolicy}$ function from Alg. \ref{algo:adaptivecompression} such that CFs are scaled up in exponents of 2 w.r.t the first initialized $\theta_{min}$.
On top of DGC, we set $\theta_{min}$ and $\theta_{max}$ to 10$\mathsf{x}$ and 1000$\mathsf{x}$, $\mathsf{window}$=500 and $\omega$=1\%.
So we scale up by factors of $2^{1}$, $2^{2}$, $2^{4}$, $2^{8}$ w.r.t 10$\mathsf{x}$ up until 1000$\mathsf{x}$.
The candidate CFs thus evaluated in this policy are 10$\mathsf{x}$, 20$\mathsf{x}$, 40$\mathsf{x}$, 160$\mathsf{x}$ and 1000$\mathsf{x}$.
We run \emph{GraVAC} on two configuration with different thresholds on compression gain, $\epsilon$ = 0.7 and 0.9.
The lower $\epsilon$ relaxes the constraint on the gain for higher CFs to be eligible for communication, thus achieving higher compression.
A large $\epsilon$ (i.e., close to 1) allows for compression only if the compressed tensors are highly representative of the original gradients.
First, we compare these two thresholds with Dense SGD as the latter demonstrates the ideal convergence scenario.
Then, we compare \emph{GraVAC} with different compression techniques on static CFs and look at final model accuracy, communication savings and overall speedup.

\begin{figure}
\hspace*{-0.55cm}
\subfloat[Test accuracy]{\includegraphics[width=0.17\textwidth]{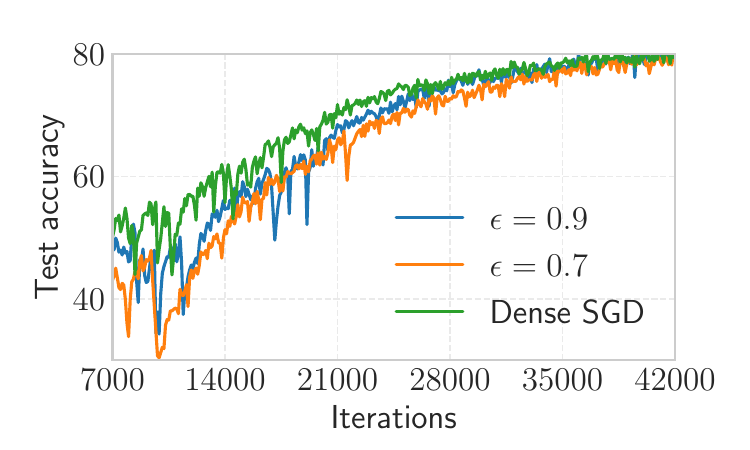}
\label{fig:resnetgputacc}}
\subfloat[Iteration density]{\includegraphics[width=0.17\textwidth]{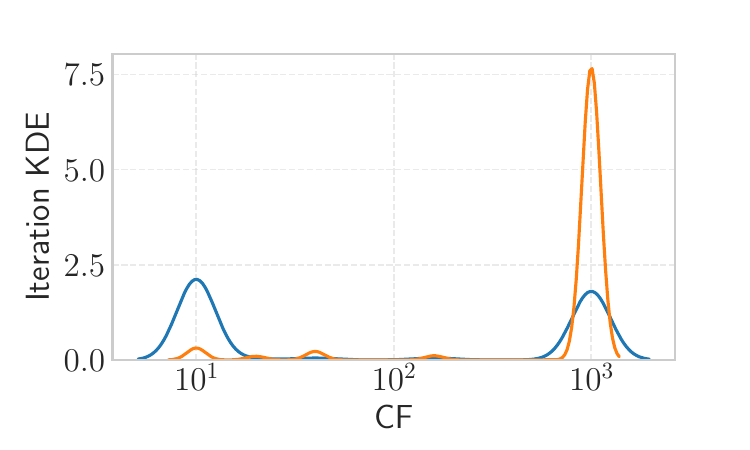}
\label{fig:resnetgputKDE}}
\subfloat[$T_{compression}$]{\includegraphics[width=0.17\textwidth]{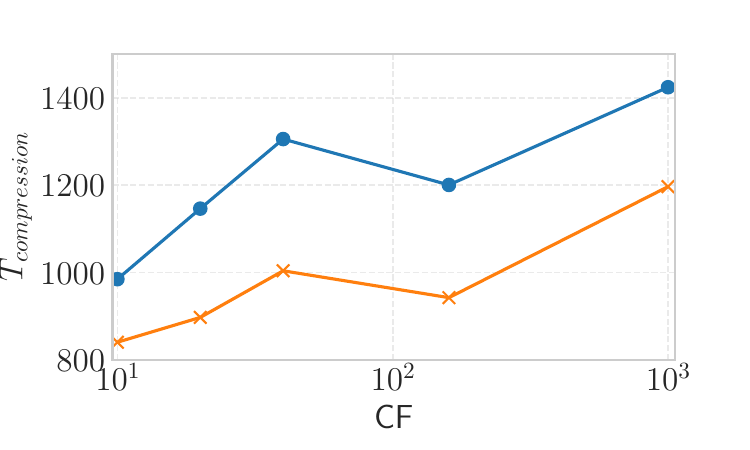}
\label{fig:resnetgputTput}}
\caption{ResNet101:\emph{GraVAC} with $\epsilon$ = [0.7, 0.9] and Dense SGD.} \label{fig:resnetgput}
\end{figure}

\emph{ResNet101:} Fig. \ref{fig:resnetgput} shows how \emph{GraVAC} achieves the same convergence as dense SGD in the same number of iterations.
The low and high $\epsilon$ reduce overall communication volume by 163$\times$ and 19$\times$ over dense SGD.
\emph{We measure communication volume as the ratio of cumulative single-precision floats exchanged among workers in GraVAC relative to dense SGD.}
Training cycle is slightly more volatile with compression, as seen from the accuracy drop due to lr decay at around 9000-th iteration.
The drop is more apparent for $\epsilon$ = 0.7 as we continue to train with higher CFs on account of the lower threshold.
Comparatively, $\epsilon$ = 0.9 is more robust to hyperparameter tuning like lr decay as we tend to train with a lower CF due to higher threshold.
This is corroborated from Fig. \ref{fig:resnetgputKDE} which shows distribution of training iterations over the CFs.
We equally train with 10$\mathsf{x}$ and 1000$\mathsf{x}$ for $\epsilon$ = 0.9, while we mostly train with 1000$\mathsf{x}$ for $\epsilon$ of 0.7.
For the compression throughputs of $\epsilon$ = 0.9 in Fig. \ref{fig:resnetgputTput}, it might seem counterintuitive at first that although $T_{compression}$ is maximum for 1000$\mathsf{x}$ and minimum for 10$\mathsf{x}$, we still evenly train with the two CFs.
This is on account of the high threshold and because $\theta_{min}$ did not scale up and remained at 10$\mathsf{x}$ for ResNet101.
Thus, whenever the compression gain of any candidate CF did not meet the threshold, we synchronized gradients compressed at 10$\mathsf{x}$.
For $\epsilon$ of 0.7, compression throughput was maximum for 1000$\mathsf{x}$ and we trained at this CF for most iterations as the corresponding gain easily met that threshold.

\begin{figure}
\hspace*{-0.55cm}
\subfloat[Test accuracy]{\includegraphics[width=0.17\textwidth]{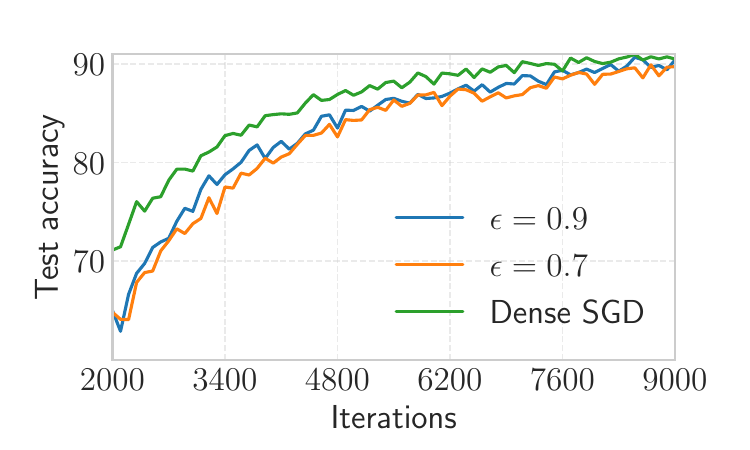}
\label{fig:vgggputacc}}
\subfloat[Iteration density]{\includegraphics[width=0.17\textwidth]{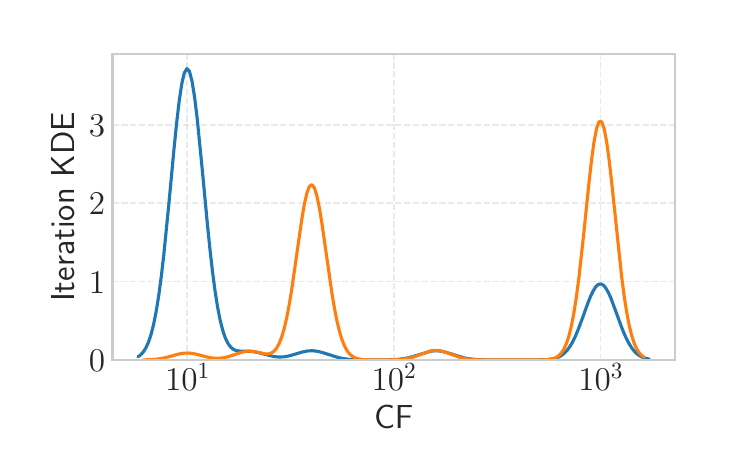}
\label{fig:vgggputKDE}}
\subfloat[$T_{compression}$]{\includegraphics[width=0.17\textwidth]{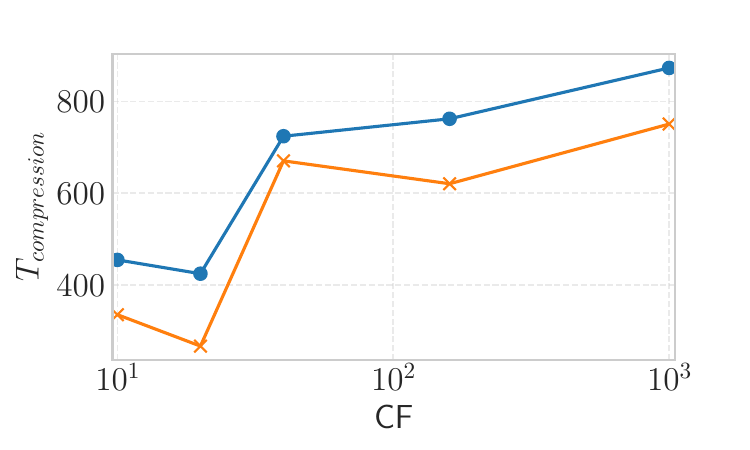}
\label{fig:vgggputTput}}
\caption{VGG16: \emph{GraVAC} with $\epsilon$ = [0.7, 0.9] and Dense SGD.} \label{fig:vgggput}
\end{figure}

\emph{VGG16:} Like ResNet101, VGG16 also converges to the same accuracy as dense SGD within the same iterations, where $\epsilon$ = 0.7 and 0.9 reduce communication volume by 80$\times$ and 13.5$\times$ over dense SGD (Fig. \ref{fig:vgggput}).
Although $T_{compression}$ is maximum at 1000$\mathsf{x}$ for $\epsilon$ = 0.9, the corresponding gain was \emph{not} as high to meet the threshold.
Because of this, we switch back to $\theta_{min}$ and thus train with 10$\mathsf{x}$ for majority iterations as seen from the kernel density estimates in Fig. \ref{fig:vgggputKDE}.
However, when $\epsilon$ was lower, we were able to find 40$\mathsf{x}$ CF to meet that threshold.
$T_{compression}$ corresponding to this CF was second largest in our exploration space.
As candidate CFs are evaluated over the iterations, the model gradually converges and as a result, compression gain improves even further on larger CFs as training progresses.
Ultimately, we arrive on $\theta_{ideal}$ = 1000$\mathsf{x}$ corresponding to the maximum compression throughput (Fig. \ref{fig:vgggputTput}).

\begin{figure}
\hspace*{-0.55cm}
\subfloat[Test accuracy]{\includegraphics[width=0.17\textwidth]{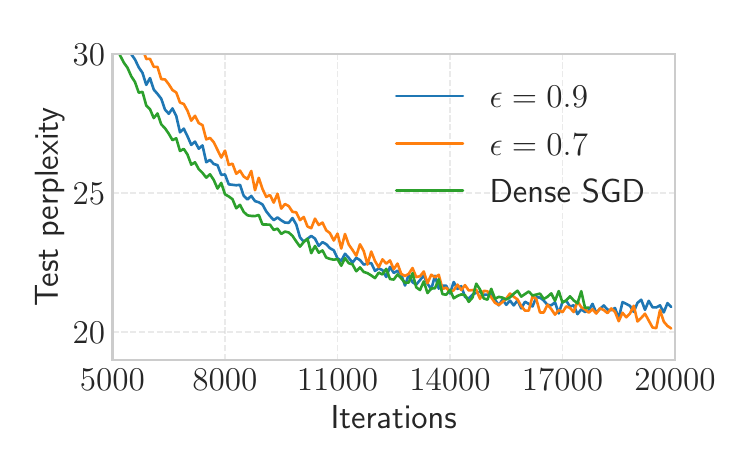}
\label{fig:lstmgputacc}}
\subfloat[Iteration density]{\includegraphics[width=0.17\textwidth]{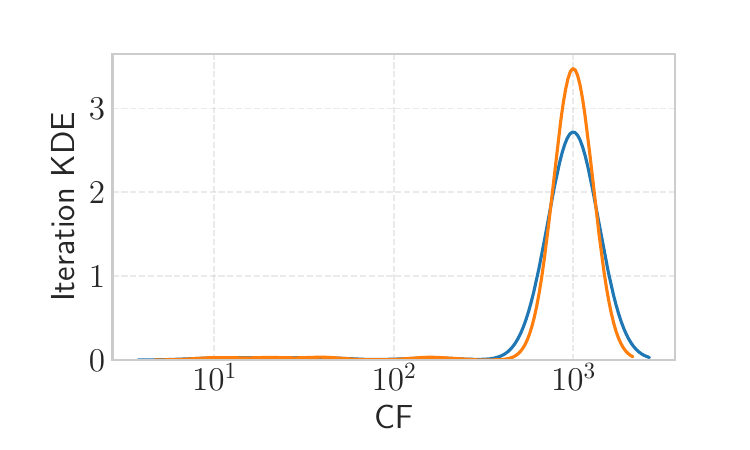}
\label{fig:lstmgputKDE}}
\subfloat[$T_{compression}$]{\includegraphics[width=0.17\textwidth]{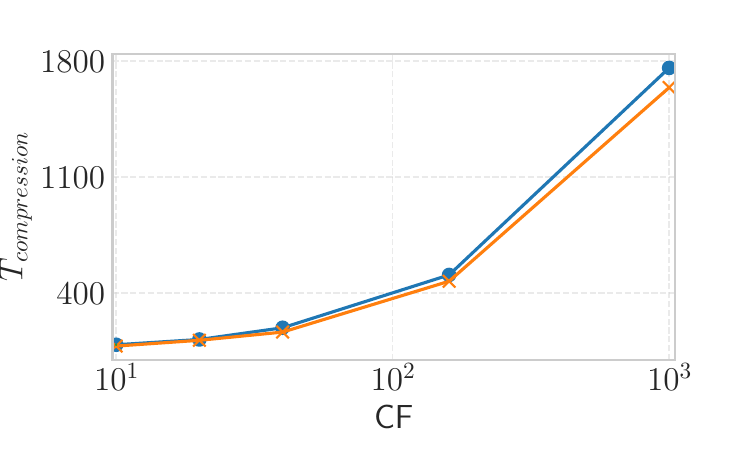}
\label{fig:lstmgputTput}}
\caption{LSTM: \emph{GraVAC} with $\epsilon$ = [0.7, 0.9] and Dense SGD.} \label{fig:vgggput}
\end{figure}

\emph{LSTM:} Like the models before, \emph{GraVAC} with either $\epsilon$ converged in the same iterations as dense SGD training, while reducing the communication volume by 279$\times$ and 289$\times$ for $\epsilon$ of 0.9 and 0.7 respectively.
Given the dataset, model and training hyperparameters, we already saw from Fig. \ref{fig:lstmcompgain} that compression gain for LSTM was high for both 10$\mathsf{x}$ and 1000$\mathsf{x}$.
We observed a similar trend here as compression gain corresponding to 1000$\mathsf{x}$ easily satisfied both thresholds and thus, we train with the largest available CF for most iterations (Fig. \ref{fig:lstmgputKDE}).
Correspondingly, the compression throughput is maximum at this CF as well.

\begin{table}[!t]
	\renewcommand{\arraystretch}{1.3}
	\caption{\emph{GraVAC}'s model quality and speedup over static CFs}
	\centering
	\begin{tabular}{|c|c|c|c|c|}
		\hline
		\bfseries Model & \bfseries Compression & \bfseries Acc./Ppl & \bfseries Diff. & \bfseries Speedup \\
		\hline
		\multirow{9}{*}{ResNet101} & Top-\textit{k} 10$\mathsf{x}$ & 80.14\% & +0.14\% & 1$\times$ \\
		\cline{2-5}
		& Top-\textit{k} 1000$\mathsf{x}$ & 76.4\% & $-$3.6\% & 3.02$\times$ \\
		\cline{2-5}
		& DGC 10$\mathsf{x}$ & 80.4\% & +0.4\% & 1.23$\times$ \\
		\cline{2-5}
		& DGC 1000$\mathsf{x}$ & 78.6\% & $-$1.4\% & 5.19$\times$ \\
		\cline{2-5}
		& Redsync 10$\mathsf{x}$ & 79.4\% & $-$0.6\% & 1.2$\times$ \\
		\cline{2-5}
		& Redsync 1000$\mathsf{x}$ & 77.4\% & $-$2.6\% & 6.94$\times$ \\
		\cline{2-5}
		& Random-\textit{k} 10$\mathsf{x}$ & - & - & - \\
		\cline{2-5}
		& Random-\textit{k} 1000$\mathsf{x}$ & - & - & - \\
		\cline{2-5}
		& \emph{\textbf{GraVAC}} & \textbf{80.2\%} & \textbf{+0.2\%} & \textbf{4.32$\times$} \\
		\hline
		\multirow{9}{*}{VGG16} & Top-\textit{k} 10$\mathsf{x}$ & 91.2\% & +1.2\% & 1$\times$ \\
		\cline{2-5}
		& Top-\textit{k} 1000$\mathsf{x}$ & 90.68\% & +0.68\% & 3.22$\times$ \\
		\cline{2-5}
		& DGC 10$\mathsf{x}$ & 90.8\% & +0.8\% & 0.935$\times$ \\
		\cline{2-5}
		& DGC 1000$\mathsf{x}$ & 90.4\% & +0.4\% & 3.35$\times$ \\
		\cline{2-5}
		& Redsync 10$\mathsf{x}$ & 90.45\% & +0.45\% & 0.99$\times$ \\
		\cline{2-5}
		& Redsync 1000$\mathsf{x}$ & 90.3\% & +0.3\% & 3.6$\times$ \\
		\cline{2-5}
		& Random-\textit{k} 10$\mathsf{x}$ & 87.8\% & $-$2.2\% & 0.7$\times$ \\
		\cline{2-5}
		& Random-\textit{k} 1000$\mathsf{x}$ & - & - & - \\
		\cline{2-5}
		& \emph{\textbf{GraVAC}} & \textbf{90.48\%} & \textbf{+0.48\%} & \textbf{1.95$\times$} \\
		\hline
		\multirow{9}{*}{LSTM} & Top-\textit{k} 10$\mathsf{x}$ & 22.0 & +0.0 & 1$\times$ \\
		\cline{2-5}
		& Top-\textit{k} 1000$\mathsf{x}$ & 26.78 & $-$4.78 & 3.36$\times$ \\
		\cline{2-5}
		& DGC 10$\mathsf{x}$ & 21.67 & +0.33 & 1.23$\times$ \\
		\cline{2-5}
		& DGC 1000$\mathsf{x}$ & 25.14 & $-$3.14 & 6.25$\times$ \\
		\cline{2-5}
		& Redsync 10$\mathsf{x}$ & 21.65 & +0.35 & 1.17$\times$ \\
		\cline{2-5}
		& Redsync 1000$\mathsf{x}$ & 24.24 & $-$2.24 & 6.9$\times$ \\
		\cline{2-5}
		& Random-\textit{k} 10$\mathsf{x}$ & 24.15 & $-$2.15 & 1.3$\times$ \\
		\cline{2-5}
		& Random-\textit{k} 1000$\mathsf{x}$ & - & - & - \\
		\cline{2-5}
		& \emph{\textbf{GraVAC}} & \textbf{21.25} & \textbf{+0.75} & \textbf{6.67$\times$} \\
		\hline
		
	\end{tabular}
	\label{table:gravacspeedup}
\end{table}

Further, we compare \emph{GraVAC} with static CFs running on different compression techniques.
In particular, we train our models with Top-\textit{k}, DGC, Redsync and Random-\textit{k} at CFs 10$\mathsf{x}$ and 1000$\mathsf{x}$.
We run each compression technique to report the final accuracy/perplexity until it does not improve any further, difference in convergence compared to dense SGD baseline from Table \ref{table:models}, and relative training speedup over Top-\textit{k} 10$\mathsf{x}$ for each model.
The results are tabulated in Table \ref{table:gravacspeedup}.
We do not consider dense SGD training in this comparison since we already established previously how \emph{GraVAC} is able to achieve the same convergence in the same iterations, and other compression techniques have already been compared to dense SGD in prior works.
For ResNet101, 1000$\mathsf{x}$ CF on Redsync, DGC and Top-\textit{k} have considerably high speedups than 10$\mathsf{x}$ Top-\textit{k}.
However, these methods at 1000$\mathsf{x}$ CF achieve considerably less accuracy than Top-\textit{k} at 10$\mathsf{x}$.
At 1000$\mathsf{x}$, Top-\textit{k}, DGC and Redsync do not improve beyond 76.4\%, 78.6\% and 77.4\% top-1 test accuracy.
Random-\textit{k} faild to converge at either CF and accuracy did not improve beyond 20\% .
Because of \emph{GraVAC}'s adaptive scheme, we converge to 80.2\% accuracy while still reducing training time by 4.32$\times$.

For VGG16, we previously observed that the model is already quite robust to high compression (Fig. \ref{fig:vggpriorpostcomp}).
We see that again here for Top-\textit{k}, DGC and Redsync at 1000$\mathsf{x}$ cross 90\% accuracy with 3.22, 3.35 and 3.6$\times$ speedup over Top-\textit{k} 10$\mathsf{x}$.
Random-\textit{k} at 10$\mathsf{x}$ also converged, albeit to a lower 87.8\% accuracy and slower convergence.
Since \emph{GraVAC} attains 90.48\% test accuracy with 1.95$\times$ training speedup, other compression schemes were more optimal in this case simply because they used high CFs.

In LSTM, \emph{GraVAC} obtains the least perplexity of 21.25 while still providing maximum speedup of 6.67$\times$ over Top-\textit{k} 10$\mathsf{x}$.
Random-\textit{k} 10$\mathsf{x}$ converged to 24.15 perplexity and did not improve further, while Random-\textit{k} 1000$\mathsf{x}$ failed here again.
Of all the configurations, only Top-\textit{k}, DGC and Redsync at 10$\mathsf{x}$ CF and \emph{GraVAC} achieved better perplexity than dense SGD.

Thus, we see how \emph{GraVAC} is able to train models like ResNet101 and LSTM to high accuracy/perplexity and still reduce training time significantly.
Static compression schemes achieve high accuracy at low CF at the cost of high communication overhead, thus providing lower speedup.
Large CFs considerably reduce communication, but the final model quality is not at par with \emph{GraVAC}.
On the flip side, some over-parameterized models like VGG16 can be robust to compression and still converge successfully at high static CFs.

\vspace{0.2cm}
\subsubsection{Geometric scaling policy}

\begin{figure}
\hspace*{-0.6cm}
\subfloat[Convergence]{\includegraphics[width=0.25\textwidth]{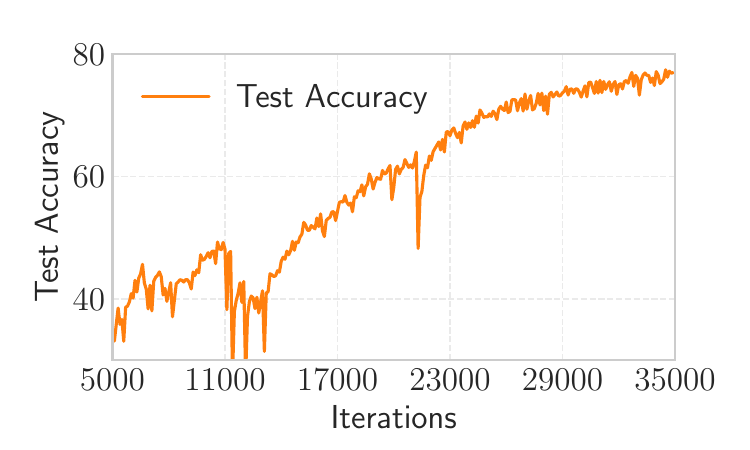}
\label{fig:resnetgradualacc}}
\subfloat[$T_{compression}$ and KDE]{\includegraphics[width=0.25\textwidth]{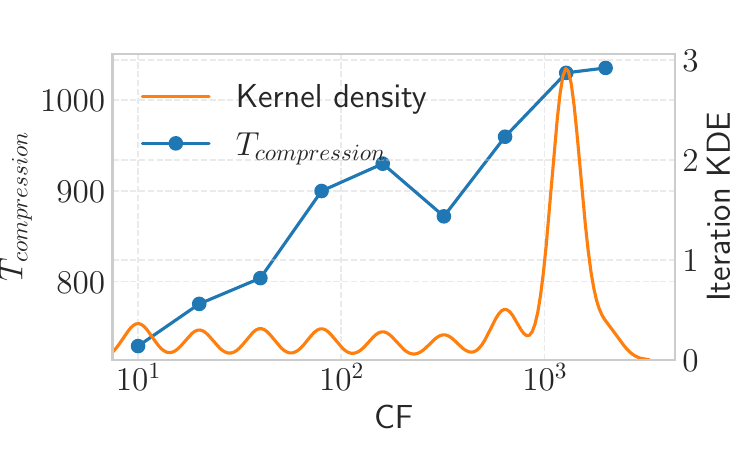}
\label{fig:resnetgradualkde}}
\caption{ResNet101: \emph{GraVAC} with Geometric scaling policy.} \label{fig:resnetgradualscaling}
\end{figure}

We also propose a relatively smoother compression policy where $\textsf{ScalingPolicy}$ increments CFs as a geometric progression with common ratio 2.
We deploy \emph{GraVAC} with Redsync on ResNet101 and set $\theta_{min}$ = 10$\mathsf{x}$, $\theta_{max}$ = 2000$\mathsf{x}$, $\epsilon$ = 0.7, $\mathsf{window}$ = 2000 steps and $\omega$ = 1\%.
Thus, candidate CFs are 10$\mathsf{x}$, 20$\mathsf{x}$, 40$\mathsf{x}$, 80$\mathsf{x}$, 160$\mathsf{x}$, 320$\mathsf{x}$, 640$\mathsf{x}$, 1280$\mathsf{x}$ and 2000$\mathsf{x}$.
Fig. \ref{fig:resnetgradualacc} shows the accuracy curve over the iterations.
Compared to dense SGD (Fig. \ref{fig:resnetgputacc}), \emph{GraVAC} with geometric scaling converged \emph{while reducing communication volume by 76$\times$}.
In contrast to exponential scaling, convergence is relatively slower because we evaluate each candidate CF for a larger $\mathsf{window}$ size.
As a result, gradients get even smaller as \emph{GraVAC} gradually arrives at larger CFs and compression gain increases beyond $\epsilon$.
Thus, we see similar iteration densities from CF 10$\mathsf{x}$ to 640$\mathsf{x}$ (Fig. \ref{fig:resnetgradualkde}).
After the first 7 CFs are evaluated over 2000 steps each, we mostly train with CF 1280$\mathsf{x}$ from 16K iterations onward (because 8 $\times$ 2000 = 16000).
We did not scale to 2000$\mathsf{x}$ in our evaluation since compression throughput for 1280$\mathsf{x}$ and 2000$\mathsf{x}$ was 1029.9 and 1035.4, which falls within $\omega$'s bound of 1\%.
\emph{This case highlights the effectiveness of GraVAC such that it does not scale the CF beyond a point when it stop improving the parallel or statistical efficiency of gradient compression}.
In this case, \emph{GraVAC} does not compress beyond 1280$\mathsf{x}$ as it corresponds to the maximum compression throughput (and at a lower CF of 1280$\mathsf{x}$ compared to 2000$\mathsf{x}$).

\subsection{Gains of Multi-level Compression in \emph{GraVAC}}\label{subsec:multilevelgravac}

Alg. \ref{algo:adaptivecompression} explains how at each iteration, \emph{GraVAC} scales compression from initial $\theta_{min}$ to current CF being evaluated (i.e., $\theta_{c}$), up to the maximum allowed $\theta_{max}$.
Thus, compressing the original gradients (computed over backward pass) twice; i.e., once over $\theta_{min}$ and then again on $\theta_{c}$ can incur significant overhead, especially on larger models.
The latency of a compressor may vary with the size of the tensor to compress as well as the target CF.
To reduce the cumulative overhead of compressing original tensors multiple times, we apply a multi-level compression scheme as follows: given a compressor $\mathcal{C}$ and tensor $\mathcal{X}$ to be compressed to CFs $\theta_{1}$ and $\theta_{2}$ such that $\theta_{2} > \theta_{1}$, rather than compressing each CF on $\mathcal{X}$ as: $$\mathcal{X}_{1} = \mathcal{C}(\theta_{1}, \mathcal{X}) \; \text{and} \; \mathcal{X}_{2} = \mathcal{C}(\theta_{2}, \mathcal{X})$$
to produce compressed tensors where $|\mathcal{X}_{2}| < |\mathcal{X}_{1}| < |\mathcal{X}|$.
In \emph{GraVAC}, we first compute $\mathcal{X}_{1}$ and then compress this tensor to $\theta_{2}^{'}$ to produce $\mathcal{X}_{2}^{'}$: $$\mathcal{X}_{1} = \mathcal{C}(\theta_{1}, \mathcal{X}) \; \Longrightarrow \mathcal{X}_{2}^{'} = \mathcal{C}(\theta_{2}^{'}, \mathcal{X}_{1}) \; : \; \theta_{2}^{'} = \frac{\theta_{2}}{\theta_{1}}$$
The resulting tensor $\mathcal{X}_{2}^{'}$ is such that $\mathcal{X}_{2}^{'} = \mathcal{X}_{2}$ for $\theta_{2}^{'} = \theta_{2}/\theta_{1}$.
The appeal of doing so is that the second compression operation is applied on a smaller tensor $\mathcal{X}_{1}$ instead of $\mathcal{X}$ again.
We tabulate the savings of multi-level compression in Table \ref{table:compressoverhead}.
Let's consider a scaling case of \emph{GraVAC} where $\theta_{min}=10\mathsf{x}$ and current CF evaluated is 1000$\mathsf{x}$.
Then multilevel \emph{GraVAC} first compresses to 10$\mathsf{x}$ and then further compresses the reduced tensors to 100$\mathsf{x}$, i.e., $\theta_{1} = 10\mathsf{x}$ and $\theta_{2}^{'} = 100\mathsf{x}$ so that $\theta_{2} = 1000\mathsf{x}$.
In direct approach, we first compress original gradients to 10$\mathsf{x}$, then compress the original gradients again to 1000$\mathsf{x}$.
From our results, we see that multi-level compression is at least 1.1$\times$ and up to 1.83$\times$ faster than directly compressing the original tensors twice.

\begin{table}[!t]
	\renewcommand{\arraystretch}{1.3}
	\caption{\emph{GraVAC}'s mulit-level (MTL) compression speedup}
	\centering
	\begin{tabular}{|c|c|c|c|c|}
		\hline
		\bfseries Model & \bfseries Method & \bfseries Direct (ms) & \bfseries MTL (ms) & \bfseries Speedup \\
		\hline
		\multirow{4}{*}{ResNet101} & Top-\textit{k} & 606 & 332 & 1.83$\times$ \\
		\cline{2-5}
		& DGC & 90 & 59 & 1.52$\times$ \\
		\cline{2-5}
		& Redsync & 33 & 29.8 & 1.1$\times$ \\
		\cline{2-5}
		& Random-\textit{k} & 23 & 14 & 1.64$\times$ \\
		\hline
		\multirow{4}{*}{VGG16} & Top-\textit{k} & 181 & 121 & 1.49$\times$ \\
		\cline{2-5}
		& DGC & 122 & 95.5 & 1.27$\times$ \\
		\cline{2-5}
		& Redsync & 101.4 & 87.7 & 1.16$\times$ \\
		\cline{2-5}
		& Random-\textit{k} & 41.6 & 31 & 1.34$\times$ \\
		\hline
		\multirow{4}{*}{LSTM} & Top-\textit{k} & 200 & 126 & 1.59$\times$ \\
		\cline{2-5}
		& DGC & 88 & 63 & 1.4$\times$ \\
		\cline{2-5}
		& Redsync & 69.4 & 46.4 & 1.5$\times$ \\
		\cline{2-5}
		& Random-\textit{k} & 56.4 & 37.4 & 1.5$\times$ \\
		\hline
	\end{tabular}
	\label{table:compressoverhead}
\end{table}

\subsection{Comparing \emph{GraVAC} with Prior Art}

In this section, we compare \emph{GraVAC} with another adaptive scheme called Accordion \cite{b30}.
For the three models, we use bounds of Rank-1 and Rank-4 for compression in Accordion, as described in \cite{b30} and compare with \emph{GraVAC} in terms of communication and time savings (i.e., training speedup) to achieve the same test accuracy/perplexity.
The savings are normalized by Accordion's performance for each respective model, shown in Table \ref{table:gravacaccordion}.
For ResNet101, \emph{GraVAC} reduces total communication volume by 44.5$\times$ and reduces training time by 1.94$\times$ over Accordion.
\emph{GraVAC} speeds up training by 5.63$\times$ over Accordion for communication-heavy models like VGG16.
In LSTM training, \emph{GraVAC} converges twice as fast by reducing communication volume up to 104.2$\times$.

Accordion is based on detecting critical regions during training, i.e., when inter-iteration gradients computed in backward pass change significantly and cross a certain user-defined threshold.
Accordion switches between 2 compression factors such that it uses the low CF in critical regions and the higher CF otherwise.
On the other hand, \emph{GraVAC} looks at information loss on account of compression (i.e., statistical efficiency) and not just relative gradient change in sensitive regions of training.
That is, \emph{GraVAC} looks at intra-iterations gradients as well (between original and gradients compressed at different CFs).
Additionally, \emph{GraVAC} scales compression across a wider range and carefully inspects intermediary CFs as potential compression candidates.
Thus, we obtain higher speedups when training with \emph{GraVAC}.

\begin{table}[!t]
	\renewcommand{\arraystretch}{1.3}
	\caption{\emph{GraVAC} vs. Accordion: Communication and Time savings}
	\centering
	\begin{tabular}{|c|c|c|c|c|}
		\hline
		\bfseries Model & \bfseries Method & \bfseries Floats sent & \bfseries Comm. sav. & \bfseries Time sav. \\
		\hline
		\multirow{2}{*}{ResNet101} & Accordion & 4.17 $\times 10^{11}$ & 1$\times$ & 1$\times$ \\
		\cline{2-5}
		& \emph{\textbf{GraVAC}} & $\mathbf{9.38 \times 10^{9}}$ & $\mathbf{44.5 \times}$ & $\mathbf{1.94 \times}$ \\
		\hline
		\multirow{2}{*}{VGG16} & Accordion & 3.83 $\times 10^{11}$ & 1$\times$ & 1$\times$ \\
		\cline{2-5}
		& \emph{\textbf{GraVAC}} & $\mathbf{1.7 \times 10^{10}}$ & $\mathbf{22.4 \times}$ & $\mathbf{5.63 \times}$ \\
		\hline
		\multirow{2}{*}{LSTM} & Accordion & 4.2 $\times 10^{11}$ & 1$\times$ & 1$\times$ \\
		\cline{2-5}
		& \emph{\textbf{GraVAC}} & $\mathbf{4 \times 10^{9}}$ & $\mathbf{104.2 \times}$ & $\mathbf{2.06 \times}$ \\
		\hline
	\end{tabular}
	\label{table:gravacaccordion}
\end{table}

\subsubsection{\emph{GraVAC} vs. Accordion on Random-\textit{k} Compression} 

\begin{figure}
\hspace*{-0.5cm}
\subfloat[Model convergence]{\includegraphics[width=0.25\textwidth]{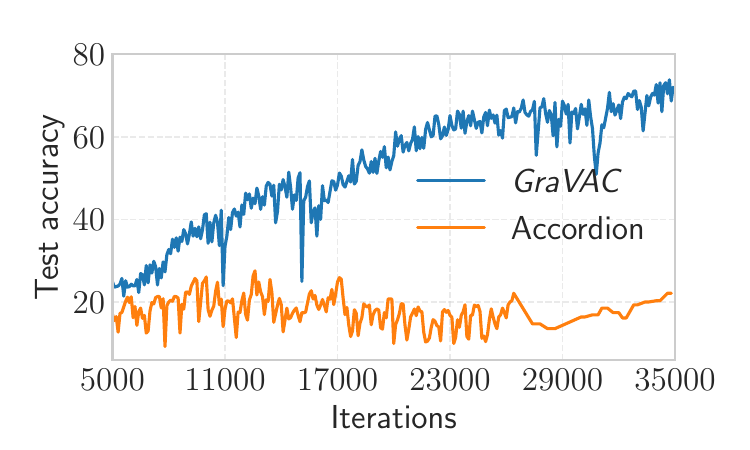}
\label{fig:randomkgravacaccordian}}
\subfloat[\emph{GraVAC} $T_{comp.}$ and KDE]{\includegraphics[width=0.25\textwidth]{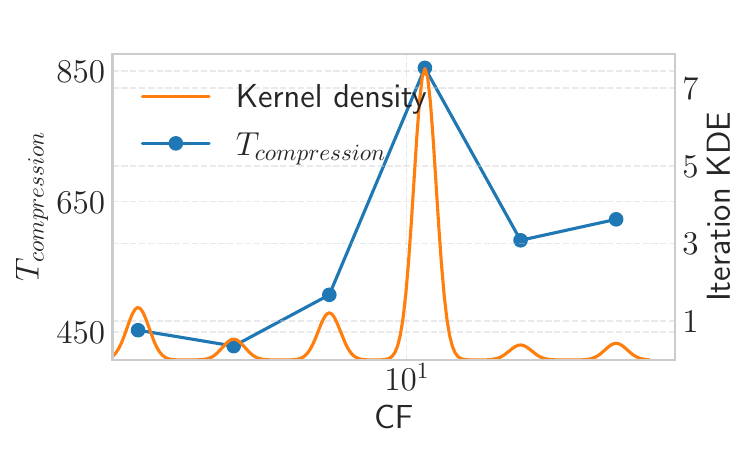}
\label{fig:gputrandomkkde}}
\caption{\emph{GraVAC} and Accordion on Random-\textit{k} compression.} \label{fig:resnetgradualscaling}
\end{figure}

We previously saw in Fig. \ref{fig:randomkbaseline} and Table \ref{table:gravacspeedup} that ResNet101 failed to converge at any CF with Random-\textit{k} compression.
In this section, we present a special case of using Random-\textit{k} under the hood with both \emph{GraVAC} and Accordion.
Although the compression quality of Random-\textit{k} is lower compared to other compressors, we present this as a special case to demonstrate how \emph{GraVAC} is more dynamic and operates at a finer granularity.
We launch \emph{GraVAC} with Random-\textit{k} on $\theta_{min}$ = 1.5$\mathsf{x}$, $\theta_{max}$ = 1000$\mathsf{x}$, $\mathsf{window}$ = 2000 and $\epsilon$ = 0.7.
The CFs are scaled up via \emph{geometric scaling policy}.
Accordion was also deployed with the same min-max bounds on CF as \emph{GraVAC}, i.e., low CF = 1.5$\mathsf{x}$ and high CF = 1000$\mathsf{x}$.
The convergence curves comparing \emph{GraVAC} and Accordion are shown in Fig. \ref{fig:randomkgravacaccordian}.
Unlike static 10$\mathsf{x}$ Random-\textit{k} compression (Fig. \ref{fig:randomkbaseline}) that failed to converge, we were able to achieve to 78\% top-1 test accuracy for ResNet101 with \emph{GraVAC}.
The CFs used for training by \emph{GraVAC} were 1.5$\mathsf{x}$, 3$\mathsf{x}$, 6$\mathsf{x}$, 12$\mathsf{x}$, 24$\mathsf{x}$ and 48$\mathsf{x}$.
All candidate CFs beyond this were ignored as they did not meet the required threshold of $\epsilon$.
CF 12$\mathsf{x}$ has the highest density, implying most iterations used this CF for training (Fig. \ref{fig:gputrandomkkde}).
Correspondingly, compression throughput is maximum for this CF as well.
Compared to dense SGD, we reduced overall communication volume by 18$\times$.

As for Accordion on Random-\textit{k}, we see in Fig. \ref{fig:randomkgravacaccordian} that training saturates at 20\% accuracy.
This is because Accordion does \emph{not} consider the efficacy of the compression technique itself, and only switches between a low and high CF if the uncompressed, inter-iteration gradients change beyond a certain measure.
With a low CF 1.5$\mathsf{x}$, information loss in Random-\textit{k} was too high to update ResNet101 in a meaningful way.

%% file: conclusion.tex
\section{Conclusion}\label{sec:conclusion}

Gradient noise has previously been used as a scalability indicator for batch and cluster-size scaling in deep learning \cite{b37, b19, b18, b35, b36}.
Adaptive compression schemes like Accordion \cite{b30} switch between two compression levels based on when the inter-iteration gradients change by some margin.
\emph{GraVAC}'s key insight is to tweak compression factor over the course of training while balancing the pareto-relationship between parallel and statistical efficiency in gradient compression.
We use ``compression gain`` to measure information loss on account of compression and choose a CF appropriately.
In our evaluation, we see that \emph{GraVAC} converges 1.95 to 6.67$\times$ faster than choosing a static CF, while converging in the same number of iterations as dense SGD.
Compared to Accordion, we observed up to 5.63$\times$ reduction in end-to-end training time.

One should be mindful when training models with \emph{GraVAC} as it introduces parameters like compression threshold ($\epsilon$) and $\mathsf{window}$ size that may affect overall training performance.
Setting too small a $\mathsf{window}$ size may result in poor convergence as all the candidate CFs may be exhausted while the model is still in early training stages and gradients are still volatile.
As for $\epsilon$, choosing a very small threshold may enable high compression but may lead to model degradation by allowing high CF gradients from the beginning that will not update the model in a significant way.